\documentclass[conference]{IEEEtran}
\usepackage{times}
\usepackage{xcolor}
\usepackage{amsfonts}
\usepackage{gensymb}
\usepackage{float}
\usepackage{subfigure}
\usepackage{wrapfig}

\usepackage{amsmath}
\DeclareMathOperator*{\argmax}{arg\,max}

\newcommand\figref {Figure~\ref}
\newcommand{\eref}{Equation~\eqref}
\usepackage{multicol}
\usepackage{amsmath}
\usepackage{amssymb}
\usepackage{booktabs}
\usepackage{multirow}
\usepackage{graphicx}
\usepackage{paralist}
\usepackage[font=scriptsize,labelfont=sl,labelsep=period]{caption}
\usepackage{subfiles}
\usepackage{float}

\newcommand{\Rot}{\mathrm{Rot}}
\newcommand{\SOtwo}{\mathrm{SO}(2)}
\newcommand{\reg}{\mathrm{reg}}
\newcommand{\quot}{\mathrm{quot}}
\newtheorem{proposition}{Proposition}

\newcommand{\SE}{\mathrm{SE}}
\newcommand{\SO}{\mathrm{SO}}
\newcommand{\pick}{\mathrm{pick}}
\newcommand{\place}{\mathrm{place}}

\usepackage[numbers]{natbib}

\usepackage{hyperref}
\hypersetup{hidelinks}

\newtheorem{theorem}{Theorem}[section]
\newtheorem{lemma}[theorem]{Lemma}

\begin{document}


% \title{Equivariant Transporter Networks for 2D Manipulation}

% \title{Equivariant Transporter Network for Pick and Place}

\title{Equivariant Transporter Network}

% You will get a Paper-ID when submitting a pdf file to the conference system
\author{Paper-ID [4]}
\author{\authorblockN{Haojie Huang\qquad Dian Wang \qquad Robin Walters\qquad Robert Platt\\}
\authorblockA{Khoury College of Computer Science\\
Northeastern University\\
Boston, MA 02115\\
\texttt \{huang.haoj; wang.dian; r.walters\} @northeastern.edu; rplatt@ccs.neu.edu}}
%\author{\authorblockN{Michael Shell}
%\authorblockA{School of Electrical and\\Computer Engineering\\
%Georgia Institute of Technology\\
%Atlanta, Georgia 30332--0250\\
%Email: mshell@ece.gatech.edu}
%\and
%\authorblockA{Twentieth Century Fox\\
%\and
%\authorblockN{James Kirk\\ and Montgomery Scott}
%\authorblockA{Starfleet Academy\\
%San Francisco, California 96678-2391\\

% avoiding spaces at the end of the author lines is not a problem with
% conference papers because we don't use \thanks or \IEEEmembership

%Homer Simpson\authorrefmark{2},
%Montgomery Scott\authorrefmark{3} and
%Eldon Tyrell\authorrefmark{4}}
%Telephone: (800) 555--1212, Fax: (888) 555--1212}
%\authorblockA{\authorrefmark{4}Tyrell Inc., 123 Replicant Street, Los Angeles, California 90210--4321}}

\maketitle

\begin{abstract}

Transporter Net is a recently proposed framework for pick and place that is able to learn good manipulation policies from a very few expert demonstrations~\cite{zeng2020transporter}. A key reason why Transporter Net is so sample efficient is that the model incorporates rotational equivariance into the pick-conditioned place module, i.e. the model immediately generalizes learned pick-place knowledge to objects presented in different pick orientations. This paper proposes a novel version of Transporter Net that is equivariant to both pick and place orientation. As a result, our model immediately generalizes pick-place knowledge to different place orientations in addition to generalizing the pick orientation as before. Ultimately, our new model is more sample efficient and achieves better pick and place success rates than the baseline Transporter Net model.

% \textcolor{red}{Our experiments show that only with 10 expert demonstrations, Equivariant Transporter Net can achieve greater than 95\% success rate on 7/10 tasks of unseen configurations of Ravens-10 Benchmark. Finally, we augment our model with the ability to grasp using a parallel-jaw gripper rather than just a suction cup and demonstrate it on both simulation tasks and a real robot. Videos and code are available at \href{https://haojhuang.github.io/etp_page/}{our project page}.}

\end{abstract}

\IEEEpeerreviewmaketitle

\section{Introduction}

Many challenging robotic manipulation problems can be viewed through the lens of a single pick and place operation. This is the approach taken in the Transporter Network framework~\cite{zeng2020transporter} where the model first detects a task-appropriate pick position and then detects a task-appropriate place position and orientation. Since the choices of pick and place pose are conditioned on the current manipulation scene, this model can be used to express multi-step pick-place policies that solve complex tasks. An important part of the Transporter Net model is the cross convolutional layer that matches an image patch around the picked object with an appropriate place position. By performing the cross correlation between an encoding of the scene and an encoding of a stack of differently rotated image patches around the pick, this model detects the task-appropriate place pose.

\begin{figure}[b]
    \centering
    \includegraphics[clip,width=0.33\textwidth]{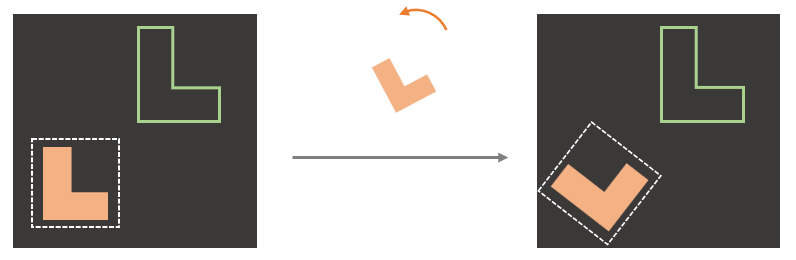}
    \caption{If Transporter Network~\cite{zeng2020transporter} learns to pick and place an object when it is presented in one orientation, the model is immediately able to generalize to new object orientations. We view this as $C_n$-equivariace of the model.}
    \label{fig:transporter_symmetry}
\end{figure}

 From the rotation-equivariance perspective that is not shown in their work, Transporter Net is equivariant with respect to pick orientation. That is, if the model can pick and place an object correctly when the object is presented in one orientation, it is automatically able to pick and place the same object when it is presented in a different orientation. This is illustrated in Figure~\ref{fig:transporter_symmetry}. The left side of Figure~\ref{fig:transporter_symmetry} shows a pick/place problem where the robot must pick the pink object and place it inside the green outline. Because the model is equivariant, the ability to solve the pick/place task on the left side of Figure~\ref{fig:transporter_symmetry} immediately implies an ability to solve the task on the right side of Figure~\ref{fig:transporter_symmetry} where the object to be picked has been rotated. This symmetry over object orientation enables Transporter Net to generalize well and it is fundamentally linked to the sample efficiency of the model. Assuming that pick orientation is discretized into $n$ possible gripper rotations, we will refer to this as a $C_n$ pick symmetry, where $C_n$ is the finite cyclic subgroup of $\SO(2)$ that denotes a set of $n$ rotations.

\begin{figure}[b]
    \centering
    \includegraphics[clip,width=0.33\textwidth]{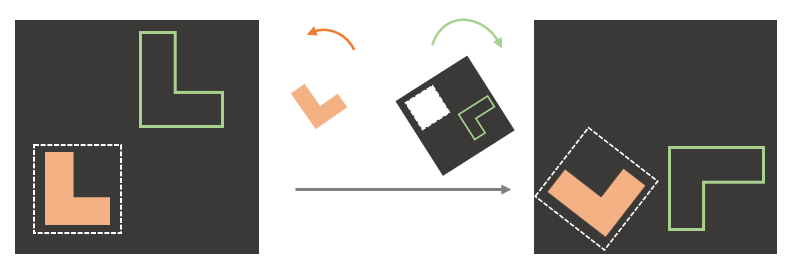}
    \caption{Our proposed Equivariant Transporter Network is able to generalize over both pick and place orientation. We view this as $C_n \times C_n$-equivariace of the model.}
    \label{fig:place_symmetry}
\end{figure}

Although Transporter Net is $C_n$-equivariant with regard to pick, the model does not have a similar equivariance with regard to place. That is, if the model learns how to place an object in one orientation, that knowledge does not generalize immediately to different place orientations. This paper seeks to add this type of equivariance to the Transporter Network model by incorporating $C_n$-equivarant convolutional layers into both the pick and place models. Our resulting model is equivariant both to changes in pick object orientation and changes in place orientation. This symmetry is illustrated in Figure~\ref{fig:place_symmetry} and can be viewed as a direct product of two cyclic groups, $C_n \times C_n$. Enforcing equivariance with respect to a larger symmetry group than Transporter Net leads to even greater sample efficiency since equivariant neural networks learn effectively on a lower dimensional action space, the equivalence classes of samples under the group action.

Our specific contributions are as follows. 1) We propose a novel version of Transporter Net that is equivariant over $C_n \times C_n$ rather than just $C_n$ and evaluate it on the Raven-10 benchmark proposed in~\cite{zeng2020transporter}. 2) We augment our Equivariant Transporter Net model with the ability to grasp using a gripper rather than just a suction cup and demonstrate it on gripper-augmented versions of five of the Ravens-10 tasks. 3) We demonstrate the approach on real-robot versions of three of the gripper-augmented tasks. Our results indicate that our approach is more sample efficient than the baseline version of Transporter Net and therefore learns better policies from a small number of demonstrations. Video and code is available at \url{https://haojhuang.github.io/etp_page/}.

% \href{https://haojhuang.github.io/etp_page/}{our project page}.}

\vspace{-0.11cm}
\section{Related Work}
\vspace{-0.11cm}
\textbf{Pick and Place.} Pick and place is an important topic in manipulation due to its value in industry. Many fundamental skills like packing, kitting, and stacking require inferring both the pick and the place action in the same time. Traditional assembly methods in factories use customized workstations so that fixed pick and place actions can be manually predefined. Recently, considerable research has focused on vision-based manipulation. Some work~\cite{narayanan2016discriminatively,chen2019grip,gualtieri2021robotic} assumes that object mesh models are available in order to run ICP~\cite{besl1992method} and align the object model with segmented observations or completions~\cite{yuan2018pcn,huang2021gascn}. Other work learns a category-level pose estimator~\cite{yoon2003real,deng2020self} or key-point detector~\cite{nagabandi2020deep,liu2020keypose} from training on a large dataset. However, these methods often require expensive object-specific labels, making them difficult to use widely. Recent advances in deep learning have provided other ways to rearrange objects from perceptual data. \citet{qureshi2021nerp} represents the scene as a graph over segmented objects to do goal-conditoned planning; \citet{curtis2021long} proposes a general system consisting of perception module, grasp module, and robot control module to solve multi-step manipulation tasks. These approaches often require a good segmentation module ahead. End-to-end models~\cite{zakka2020form2fit,khansari2020action,devin2020self,berscheid2020self} that directly map input observations to actions can learn quickly and generalize well, but most methods need to be trained on large datasets. For example, \citet{khansari2020action} collects a dataset with 7.2 million samples. \citet{devin2020self} collects $40K$ grasps and places per task. \citet{zakka2020form2fit} collects 500 disassembly sequences for each kit. Instead, our proposed method improves the sample efficiency of end-to-end models on various manipulation tasks.

\textbf{Equivariance Learning in Manipulation.}
Fully Convolutional Networks (FCN) are translationally equivariant, and have been shown to improve learning efficiency in many manipulation tasks~\cite{zeng2018robotic,morrison2018closing}. The idea of encoding SE(2) symmetries in the structure of neural networks is first introduced in G-Convolution~\cite{cohen2016group}. The extension work proposes an alternative architecture, Steerable CNN~\cite{cohen2016steerable}.~\citet{weiler2019general} proposes a general framework for implementing E(2)-Steerable CNNs. In the context of robotics learning, ~\citet{wang2022equivariant} uses SE(2) equivariance in Q learning to solve multi-step sequential manipulation tasks; \citet{wang2022mathrm} extends it to $\SOtwo$-equivairant reinforcement learning. Our work tackles manipulation rearrangement tasks by extracting inherent SE(2) equivariance through the imitation learning approach~\cite{hussein2017imitation,hester2018deep,vecerik2017leveraging}.

%(b) standard representation $\rho_1$ acts on the channel space of a vector field by rotating the vector at each pixel. One can distinguish the combination of rotations on both the spatial space and the channel space in the middle-left figure from the rotation on the spatial space only, 
% \begin{pmatrix}
%     \rightarrow & \leftarrow \\
%     \nearrow & \nwarrow
% \end{pmatrix}.\\ 

% \begin{figure}[htbp]
%     \centering
%      \centering
%      \subfloat[]{ \includegraphics[clip,width=0.07\textwidth]{figs/rho_0_.png}}
%      \hspace{0.5cm}
%      \subfloat[]{ \includegraphics[clip,width=0.15\textwidth]{figs/rho_n.png}}
%     \caption[Representation and signal transformation]{Representation and transformation of 2D signals. All types of transformations acts on the feature map by rotating the pixels and transforming the channel space. (a) trivial representation $\rho_0$ identically maps the 1-channel feature. (b) regular representation $\rho_{\mathrm{reg}}$ permutes the channel.}
%     \label{fig:signal_transformation}
% \end{figure}

% \section{Preliminaries}
\section{Background on Symmetry Groups}
\label{sect:background}

\subsection{The Groups $\SOtwo$ and $C_n$}
We are primarily interested in rotations expressed by the group $\SOtwo$ and its cyclic subgroup $C_n \leq \SOtwo$. $\SOtwo$ contains the continuous planar rotations $\{ \Rot_{\theta}: 0\leq \theta < 2\pi\}$. The discrete subgroup $C_n = \{ \Rot_{\theta}: \theta \in \{\frac{2\pi i}{n}| 0\leq i < n\} \}$ contains only rotations by angles which are multiples of $2\pi/n$. The special Euclidean group $\mathrm{SE}(2) = \SOtwo \times \mathbb{R}^2$ describes all translations and rotations of $\mathbb{R}^2$.

%For instance, the cyclic subgroup $C_4$ models the $4$ rotations which are multiples of $\frac{\pi}{2}$, i.e., $\{0,\frac{\pi}{2},{\pi},\frac{2\pi}{3}\}$.

\subsection{Representation of a Group}

A $d$-dimensional \emph{representation} $\rho \colon G \to \mathrm{GL}_d$ of a group $G$ assigns to each element $g \in G$ an invertible $d\!\times\! d$-matrix $\rho(g)$.  Different representations of $\SOtwo$ or $C_n$ help to describe how different signals are transformed under rotations.  For example, the trivial representation $\rho_0 \colon \SOtwo \to \mathrm{GL}_1$ assigns $\rho_0(g) = 1$ for all $g \in G$, i.e. no transformation under rotation. The standard representation
\[\rho_1(\Rot_\theta) = \begin{pmatrix}
\cos{\theta} & -\sin{\theta} \\
\sin{\theta} & \cos{\theta}
\end{pmatrix}
\]
represents each group element by its standard rotation matrix. Notice that $\rho_0$ and $\rho_1$ can be used to represent elements from either $\SO(2)$ or $C_n$. The regular representation $\rho_{\mathrm{reg}}$ of $C_n$ acts on a vector in $\mathbb{R}^{n}$ by cyclically permuting its coordinates $\rho_{\reg}(\Rot_{2\pi /n})(x_0,x_1,...,x_{n-2},x_{n-1})=(x_{n-1} ,x_0,x_1,...,x_{n-2})$. We can rotate by multiples of $2\pi/n$ by $\rho_{\reg}(\Rot_{2\pi i /n}) = \rho_{\reg}(\Rot_{2\pi /n})^i$. The regular representation for elements of the quotient group is denoted $\rho_{\quot}^{C_n/C_k}$ and acts on $\mathbb{R}^{n/k}$ by permuting $|C_n|/|C_k|$ channels. This gives a quotient representation of $C_n$ defined $\rho_{\quot}^{C_n/C_k}(\Rot_{2\pi i /n})(\mathbf{x})_j = (\mathbf{x})_{j + i\  \mathrm{mod} (n/k)}$, which implies features that are invariant under the action of $C_k$. For more details, we refer the reader to~\citet{serre1977linear,weiler2019general}.

% Similar to regular representation, the quotient representation of $C_n$ for $k$ dividing $n$ is denoted $\rho_{\quot}^{C_n/C_k}$ and acts on $\mathbb{R}^{n/k}$ by permutation $\rho_{\quot}^{C_n/C_k}(\Rot_{2\pi i /n})(\mathbf{x})_k = (\mathbf{x})_{k + i\  \mathrm{mod} (n/k)}$. For detailed information, we refer interested readers to~\citet{serre1977linear}.

% \subsection{Transformations of 2D Signals}
\subsection{Feature Map Transformations}

\begin{wrapfigure}[12]{r}{0.15\textwidth}
  \begin{center}
    \vspace{-1.4cm}
    \includegraphics[width=0.15\textwidth]{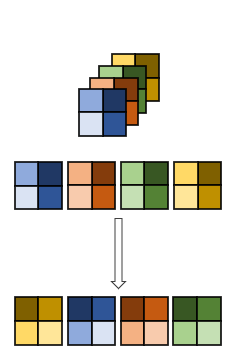}
  \end{center}
    \caption{Illustration of the action of $T_g^{\reg}$ on a $2 \times 2$ image.}
    \label{fig:rotation_illustration}
\end{wrapfigure}

We formalize images and feature maps as feature vector fields, i.e. functions $f \colon \mathbb{R}^2 \rightarrow \mathbb{R}^c$, which assign a feature vector $f(\mathbf{x}) \in \mathbb{R}^c$ to each position $\mathbf{x} \in \mathbb{R}^2$. While in practice we discretize and truncate the domain of $f$ $\lbrace (i,j) : 1 \leq i \leq W, 1 \leq j \leq W \rbrace$, here we will consider it to be continuous for the purpose of analysis. The action of a rotation $g \in \SOtwo$ on $f$ is a combination of a rotation in the domain of $f$ via $\rho_1$ (this rotates the pixel positions) and a rotation in the channel space $\mathbb{R}^c$ by $\rho \in \{\rho_0, \rho_{\reg}\}$. If $\rho = \rho_{\reg}$, then the channels cyclically permute according to the rotation. If $\rho = \rho_0$, the channels do not change. We denote this action (the action of $g$ on $f$ via $\rho$) by $T^{\rho}_g(f)$:
\begin{equation}
\label{eqn:rotation_illustration}
[T^{\rho}_g(f)](\mathbf{x}) = \rho(g) \cdot f( \rho_1(g)^{-1} \mathbf{x}).
\end{equation}
For example, the action of $T^{\rho_{\reg}}_g(f)$ is illustrated in Figure~\ref{fig:rotation_illustration} for a rotation of $g = \pi/2$ on a $2 \times 2$ image $f$ that uses $\rho_{\reg}$. The expression $\rho_1(g)^{-1} \mathbf{x}$ rotates the pixels via the standard representation. Multiplication by $\rho(g) = \rho_{\reg}(g)$ permutes the channels. For brevity, we will denote $T_g^{\reg} = T_g^{\rho_{\reg}}$ and $T_g^{0} = T_g^{\rho_{0}}$. 

% Scalar fields in which values move but are otherwise unchanged under rotation are thus transformed by  $T^{0}$.  This describes how pixel color values change when an image is rotated. 
% In contrast, $T^{\reg}$ and $T^{1}$ both rotate vector fields by both moving the location of vectors and rotating or permuting the vectors themselves.   Note that we may decompose $T^{\rho}(f) = \rho(g) \cdot  T^0_g(f)$ when convenient.
%Trivial representation could be identified as the scalar field, i.e., $f: \mathbb{R}^2 \rightarrow \mathbb{R}$; regular representation of a group $G$ is in regular feature field, i.e., $f: \mathbb{R}^2 \rightarrow \mathbb{R}^{|G|}$.

\subsection{Equivariant Mappings}

A function $F$ is equivariant if it commutes with the action of the group,
% Both the function and the neural network we use to model it may be described as respecting certain symmetry. Formally, a functional mapping $F$ is \emph{equivariant} to a group $G$ if the output produced by $F$ transforms consistently when the input transforms under the action of an element $g \in G$,
\begin{equation}
\label{eqn:equivariance}
    T_g^{\mathrm{out}}[F(f)] = F(T_g^{\mathrm{in}}[f])
\end{equation}
where $T_g^{\mathrm{in}}$ transforms the input to $F$ by the group element $g$ while $T_g^{\mathrm{out}}$ transforms the output of $F$ by $g$. For example, if $f$ is an image, then  $\SO(2)$-equivariance of $F$ implies that it acts on $f$ in the same way regardless of the orientation in which $f$ is presented. That is, if $F$ takes an image $f$ rotated by $g$ (RHS of Equation~\ref{eqn:equivariance}), then it is possible to recover the same output by evaluating $F$ for the un-rotated image $f$ and rotating its output (LHS of Equation~\ref{eqn:equivariance}).

\section{Transporter Network}

Before describing our variation on Transporter Net, we summarize the the pick-and-place problem and analyze the original Transporter Net architecture from a novel view that is not clearly illustrated in their work ~\cite{zeng2020transporter}.

% \subsection{Problem Statement}
\subsection{Problem Statement}

% Roughly following the formulation in~\cite{zeng2020transporter}, w
We define the \emph{Planar Pick and Place} problem as follows. Given a visual observation $o_t$, the problem is to learn a probability distribution $p(a_{\pick} | o_t)$ over picking actions $a_{\pick} \in \SE(2)$ and a distribution $p(a_{\place}|o_t,a_{\pick})$ over placing actions $a_{\place} \in \SE(2)$ conditioned on $a_{pick}$ that accomplishes some task of interest.
% \begin{equation}
%     f(o_t)\rightarrow p(a_{pick}),
% \end{equation}
% \begin{equation}
%     h(o_t,a_{pick})\rightarrow p(a_{place}|o_t,a_{pick}).
% \end{equation}
% In planar manipulation~\cite{levine2016end,kumra2020antipodal}, t
The visual observation $o_t$ is typically a projection of the scene (e.g., top-down RGB-D images) and the pose of the end effector is expressed as $(u,v,\theta)$ where $u,v$ denote the pixel coordinates of the gripper position and $\theta$ denotes gripper orientation. (Since~\cite{zeng2020transporter} uses suction cups to pick, that work ignores pick orientation.)

% . For a known camera pose, a bijection exists between u-v pixel coordinates and x-y Cartesian coordinates, thus transforming the task from spatial action learning to visual learning.

% , i.e., $a_{pick}$ and $a_{place}$ can also be represented by $(u,v,\theta)$. To be clear, the $\theta \in $  $a_{place}$ is the rotation from the picking pose to the placing pose, while the picking angle $\theta \in a_{pick}$ is defined as the absolute angle with respect to the world coordinate along the z-axis.

% \textcolor{red}{in supplementary part, illustrate that pushing object could be parameterized with $a_{pick}$ and $a_{place}$, starting pose of the ee and ending pose of the ee.}

\begin{figure}[htp]
    \centering
    \includegraphics[width=0.38\textwidth]{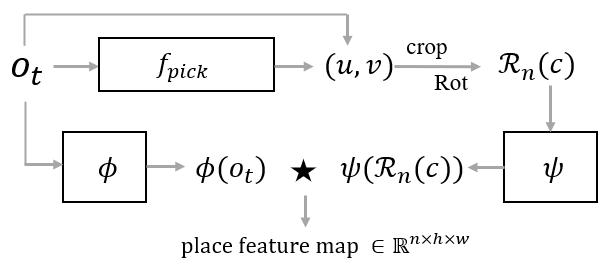}
    \caption[Transporter Network]{The Architecture of Transporter Net.}
    \label{fig:transporter_architecture}
\end{figure}

\subsection{Description of Transporter Net}
\label{sect:transporter_desc}

Transporter Network~\cite{zeng2020transporter} solves the planar pick and place problem
% . It takes a visual observation of the manipulation scene $o_t$ as input and generates a pick position $a_{pick}$ and a place pose $a_{place}$ as output.
% Transporter Network~\cite{zeng2020transporter} is an architecture for planar vision-based manipulation. It takes as input an RGB-D image $o_t\in \mathbb{R}^{4\times H\times W}$ of the scene and generates as output a pick position $(x_1,y_1)$ and a place position and orientation $(x_2,y_2,\theta_2)$. 
using the architecture shown in Figure~\ref{fig:transporter_architecture}. The pick network $f_{\pick} \colon o_t \mapsto p(u,v)$ maps $o_t$ onto a probability distribution $p(u,v)$ over pick position $(u,v) \in \mathbb{R}^2$. The output pick position $a_{\pick}^*$ is calculated by maximizing $f_{\pick}(o_t)$ over $(u,v)$. The place position and orientation is calculated as follows. First, an image patch $c$ centered on $a_{\pick}^*$ is cropped from $o_t$ to represent the pick action as well as the object. Then, the crop $c$ is rotated $n$ times to produce a stack of $n$ rotated crops. Using the notation of Section~\ref{sect:background}, we will denote this stack of crops as
\begin{equation}
    \mathcal{R}_n(c) = (T^0_{2\pi i/n}(c))_{i=0}^{n-1},
\end{equation}
where we refer to $\mathcal{R}_n$ as the ``lifting'' operator. Then, $\mathcal{R}_n(c)$ is encoded using a neural network $\psi$.
%Transporter Net also takes the original image $o_t$ and masks out the crop: $o_t \backslash c$.
The original image, $o_t$, is encoded by a separate neural network $\phi$. The distribution over place location is evaluated by taking the cross correlation between $\psi$ and $\phi$,
\begin{equation}
\label{eqn:transporter1}
    f_{\place}(o_t,c) = \psi(\mathcal{R}_n(c)) \star \phi(o_t),
\end{equation}
where $\psi$ is applied independently to each of the rotated channels in $\mathcal{R}_n(c)$.  Place position and orientation is calculated by maximizing $f_{\place}$ over the pixel position (for position) and the orientation channel (for orientation).

% As shown in~\figref{fig:euqi_transporter}, after identifying the picking pixel colored in red (by a pick network, not shown), a local image patch centered at the picking location is cropped from the top-down observation $o_t$. We will write this patch as a feature vector field $c : \mathbb{R}^2 \rightarrow \mathbb{R}^4$. This crop is rotated $n$ times (once for each element in $C_n$) to produce a stack of $n$ rotated crops around the pick location, $\mathcal{R}_n(c) = (T^0_{2\pi i/n}(c))_{i=0}^n$, and then encoded using a network $\psi$. Transporter Net detects place locations by performing a cross correlation operation between $\psi(\mathcal{R}_n(c))$ and an encoding of the original image. Let $o_t \backslash c$ denote the original observation with $c$ masked out. We encode this image using another neural network $\phi$ and then evaluate a map over place positions $Q$, by performing the cross convolutional operation, 
% \begin{equation}
% \label{eqn:transporter1}
% Q = \psi (\mathcal{R}_n(c)) \star \phi (o_t \backslash c).
% \end{equation}
%======================================%
\begin{figure}[tp]
    \centering
    \includegraphics[width=0.35\textwidth]{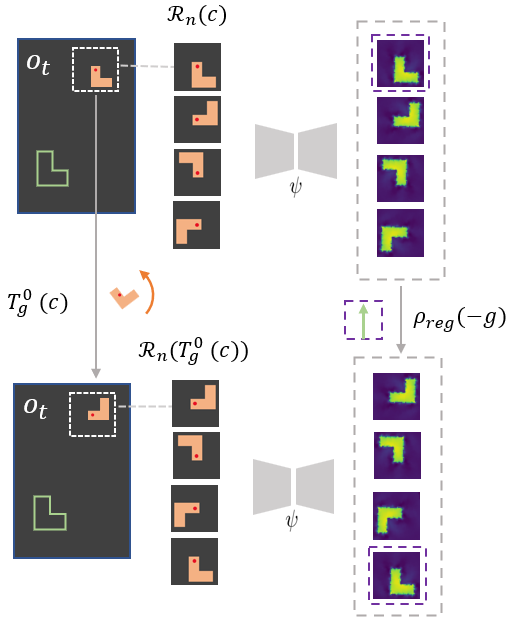}
    \caption{Illustration of the main part of the proof of Proposition~\ref{prop:equivtransporter}.   Rotating the crop $c$ induces a cyclic shift in the channels of the output $\psi(\mathcal{R}_n(T_g^0 c)) = \rho_\mathrm{reg}(-g)\psi(\mathcal{R}_n(c)).$ 
    %That is, $\psi(\mathcal{R}_n(\cdot))$ is equivariant.
    }
    % \caption[Equivariance of Transporter Network]{Equivariance of Transporter Network under the rotation of the object. A $\pi/2$ rotation on the L-shape block is equivariant to a permutation of the $4$ feature maps from $\psi$. The best matching (encompassed by the purple box) of the feature map to the L-shape slot undergoes an inverse permutation.}
    \label{fig:euqi_transporter}
    
\end{figure}

\subsection{Equivariance of Transporter Net}

The model architecture described above gives Transporter Network the following equivariance property.

\begin{proposition}
\label{prop:equivtransporter}
The Transporter Net place network $f_{\place}$ is $C_n$-equivariant. That is, given $g \in C_n$, object image crop $c$ and scene image $o_t$, 
\begin{equation}
    \label{eqn:transporter3}
    f_{\place}(o_t,T^0_g(c))  = \rho_{\reg}(-g) f_{\place}(o_t,c).
\end{equation}
% \begin{equation}
%     \label{eqn:transporter3}
%     \psi(\mathcal{R}_n[T^0_g(c)]) \star \phi (o_t \backslash c) = \rho_{reg}(-g) (\psi(\mathcal{R}_n[c]) \star \phi (o_t\backslash c)).
% \end{equation}
\end{proposition}

% \begin{proof}
% See Appendix~\ref{}.
% \end{proof}

Proposition~\ref{prop:equivtransporter} expresses the following intuition. A rotation of $g$ applied to the orientation of the object to be picked results in a $-g$ change in the placing angle, which is represented by a permutation along the channel axis of the placing feature maps. This is a symmetry over the cyclic group $C_n \leq \SO(2)$ which is encoded directly into the model. It enables it to immediately generalize over different orientations of the object to be picked and thereby improves sample efficiency.

The main idea of the proof is shown in Figure \ref{fig:euqi_transporter}.  Namely $\psi(\mathcal{R}_n(\cdot))$ is equivariant in the sense that rotating the crop $c$ induces a cyclic shift in the channels of the output. Formally, $\psi(\mathcal{R}_n(T_g^0 c)) = \rho_\mathrm{reg}(-g)\psi(\mathcal{R}_n(c)).$  Noting that a permutation of the filters $K$ in the convolution $K \star \phi(o_t)$ induces the same permutation in the output feature maps completes the proof. The full proof is given in Appendix~\ref{proof_proposition}.  Note that here $\psi$ is a simple CNN with no rotational equivariance.  The equivariance results from the lifting $\mathcal{R}_n$.

 However, only the place network of Transporter Net has the $C_n$-equivariance. Instead, our proposed method incorporates the rotational equivariance in the pick network and  $C_n\times C_n$-equivariance in the place network.

% \begin{equation}
%     \psi(\mathcal{R}_n(T^0_{g} c)) = \rho_{\reg}(g) \psi(\mathcal{R}_n(c)),
% \end{equation}

% The proof of Proposition~\ref{prop:equivtransporter} relies on equivariance properties of $\mathcal{R}_n$, $\psi$, and $\star$.  In particular, 
% \begin{equation}
% \label{eqn:transporter2}
%     \mathcal{R}_n(T^0_{g} c) = \rho_{\reg}(g) \mathcal{R}_n(c),
% \end{equation}
% for $g = 2 \pi i /n$. On the LHS, we have a stack of images $\mathcal{R}_n(T^0_{g} c)$ starting with a rotated image patch $T^0_{g} c$. The RHS encodes the recognition that this stack of patches can be calculated by producing a stack of images $\mathcal{R}_n(c)$ starting with the original patch $c$ and then cyclically shifting the patches by $i$ places where $g = 2\pi i / n$ (using the regular representation, $\rho_{\reg}$). 

\section{Equivariant Transporter}

%=======================================%
%======================================%
\subsection{Equivariant Pick}

% \begin{figure}
%     \centering
%     \includegraphics[width=0.25\textwidth]{figs/attention1.png}
%     %\setlength{\belowcaptionskip}{-0.5cm}
%     % \caption[Picking Network Architecture of Equivariant Transporter]{Picking Network Architecture of Equivariant Transporter. After identifying the best picking location $(\hat{u},\hat{v})$, a crop centered at $(\hat{u},\hat{v})$ is fed to $f_\theta$ to decode the picking angle distribution.}
%     \caption{Equivariant Transporter Pick model. First, we find the pick position $a^*_{pick}$ by evaluating the argmax over $f_p(o_t)$. Then, we evaluate $f_\theta$ for the image patch centered on $a^*_{pick}$.}
%  \label{fig:pick_archi}
% \end{figure}

Our approach to the pick network is similar to that in Transporter Net~\cite{zeng2020transporter} except that: 1) we explicitly encode equivariance constraints into the pick networks, thereby making pick learning more sample efficient; 2) we infer pick orientation so that we can use parallel jaw grippers rather than just suction grippers.

\begin{wrapfigure}[17]{r}{0.25\textwidth}
\vspace{-0.5cm}
  \begin{center}
    \includegraphics[width=0.25\textwidth]{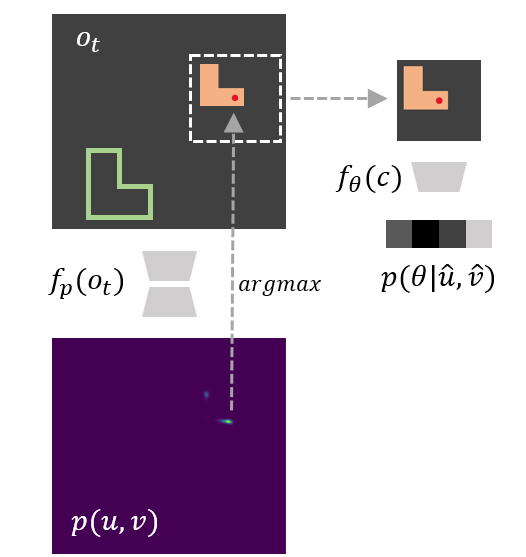}
  \end{center}
    \caption{Equivariant Transporter Pick model. First, we find the pick position $a^*_{pick}$ by evaluating the argmax over $f_p(o_t)$. Then, we evaluate $f_\theta$ for the image patch centered on $a^*_{pick}$.}
    \label{fig:pick_archi}
\end{wrapfigure}
% Considering a rotation $g\in C_n$ applied to the observation $o_t$, the picking location distribution $p(u,v)$ is desirable to undergo the same rotation without changing its value on the channel space, while the picking angle value $\theta$ will be added by $g$. When discretizing $\theta \in [0,2\pi)$ into $SO(2)$ subgroup $C_n$ to form a n-category classification problem, i.e., $\theta \in C_n$, $p(\theta)\in \mathbb{R}^n$, one may see an action $g$ on the picking angle is represented as a permutation in $p(\theta)$.

\subsubsection{Model}

We propose a $C_n$-equivariant model for detecting the planar pose for the pick operation. First, we decompose the learning process of $a_{\pick} \in \SE(2)$ into two parts,
\begin{equation}
    p(a_{\pick}) = p(u,v) p(\theta|(u,v)),
\end{equation}
where $p(u,v)$ denotes the probability that a pick exists at pixel coordinates $u,v$ and $p(\theta|(u,v))$ is the probability that the pick at $u,v$ should be executed with a gripper orientation of $\theta$. The distributions $p(u,v)$ and $p(\theta|(u,v))$ are modeled as two neural networks:
\begin{align}
\label{eqn:pick1}
f_{p}(o_t) &\mapsto p(u,v), \\
\label{eqn:pick2}
f_{\theta}(o_t,(u,v)) &\mapsto p(\theta|(u,v)).
\end{align}
% where $\mathrm{crop}(o_t,(u,v))$ denotes the image patch cropped from $o_t$ centered at $u,v$. 
Given this factorization, we can query the maximum of $p(a_{\pick})$ by evaluating $(\hat{u},\hat{v}) = \argmax_{(u,v)}[p(u,v)]$ and then $\hat{\theta} = \argmax_{\theta}[p(\theta|\hat{u},\hat{v})]$.
This is illustrated in \figref{fig:pick_archi}. The left side of \figref{fig:pick_archi} shows the maximization of $f_p$ at $a^*_{pick}$. The right side shows evaluation of $f_\theta$ for the image patch centered at $a^*_{pick}$.

% To exploit the two different types of symmetries of picking location and picking angle, we decompose the learning process of $a_{pick}=(u,v,\theta)$ into two parts:
% \begin{equation}
%     p(a_{pick}) = p(u,v)*p(\theta|(u,v))
% \end{equation}
% by two equivariant neural networks $f_p$ and $f_{\theta}$:
% \begin{equation}
%     f_{p}(o_t) \rightarrow p(u,v)
% \end{equation}
% \begin{equation}
%     f_{\theta}(o_t,(u,v)) \rightarrow p(\theta|(u,v))
% \end{equation}
% where $p(u,v)$ corresponds each pixel in $o_t$ to a picking location value, and $p(\theta|(u,v))$ represents the distribution of picking orientation conditioned on the picking location. 

% The best picking location $(\hat{u},\hat{v})$ and picking orientation $\hat{\theta}$ can be queried by:
% \begin{equation}
%     (\hat{u},\hat{v}) = \argmax_{(u,v)}[p(u,v)]
% \end{equation}
% \begin{equation}
%     \hat{\theta} = \argmax_{\theta}[p(\theta|\hat{u},\hat{v})]
% \end{equation}

\subsubsection{Equivariance Relationships}

There are two equivariance relationships that we would expect to be satisfied for planar picking:
\begin{align}
    \label{eqn:equi1}
    f_p(T^0_g(o_t)) &= T^0_g (f_p(o_t))\\
    \label{eqn:equi2}
    f_\theta(T^0_g(o_t),T^0_g(u,v)) &= \rho_{reg}(g)(f_{\theta}(o_t,(u,v))).
\end{align}
Equation~\ref{eqn:equi1} states that the grasp points found in an image rotated by $g \in \SO(2)$, (LHS of Equation~\ref{eqn:equi1}), should correspond to the grasp points found in the original image subsequently rotated by $g$, (RHS of Equation~\ref{eqn:equi1}). Equation~\ref{eqn:equi2} says that the grasp orientation at the rotated grasp point $T^0_g(u,v)$ in the rotated image $T^0_g(o_t)$ (LHS of Equation~\ref{eqn:equi2}) should be shifted by $g = 2\pi i$ relative to the grasp orientation at the original grasp points in the original image (RHS of Equation~\ref{eqn:equi2}). We encode both $f_p$ and $f_\theta$ using equivariant convolutional layers~\cite{weiler2019general} which constrain the models to represent only those functions which satisfy Equations~\ref{eqn:equi1} and~\ref{eqn:equi2}.

\subsubsection{Gripper Orientation Using the Quotient Group}

A key observation in planar picking is that, for many robots, the gripper is bilaterally symmetric, i.e. grasp outcome is invariant when the gripper is rotated by $\pi$. We can encode this additional symmetry to reduce redundancy and save computational cost using the regular representation of the quotient group $C_n / C_2$ which identifies orientations that are $\pi$ apart.  When using this quotient group for gripper orientation, $\rho_{\reg}$ in Equation~\ref{eqn:equi2} is replaced with $\rho_{\reg}^{C_n/C_2}$.

% To realize Equation \eqref{equi_pick_angle}, regular representation could be used for $p(\theta)$. But the following part provides a better choice for parallel-jaw gripper.

% With $180\degree$ symmetry of the two-finger gripper, the pick action $(x,y,\theta)$ is equivariant to $(x,y,\theta+\pi)$, and hence we shrink the picking angle $\theta$ from $C_n$ on $[0,2\pi)$ to $C_n/2$ on $[0,\pi)$. Quotient representation $\rho^{C_n/2}$ is selected to represent $p(\theta)$, which can output the same performance of regular representation but with a relatively small computation load, and thus enables a large rotation group in our picking model.

% For suction gripper that only requires 2 DoF $(x,y)$ to pick an object with suction cups, we can simply omit the part of learning $\theta$.

\subsection{Equivariant Place}

%{\color{red}RW: it would be nice if this section has a para similar to 1) above in which the symmetry is discussed independent of the network.  This makes it clear it is a symmetry of the ground truth and not the architecture. Done}

Given the picked object represented by the image patch c centered on $a_{\pick}$, the place network models the distribution of $a_{\place}=(u_{\place},v_{\place},\theta_{\place})$ by:
\begin{equation}
    f_{\place}(o_t,c) \mapsto p(a_{\place}|o_t,a_{\pick}),
\end{equation}
where $p(a_{\place}|o_t,a_{\pick})$ denotes the probability that the object at $a_{\pick}$ in scene $o_t$ should be placed at $a_{\place}$.
% the place at $u,v$ should be executed with a rotation $\theta_{\place}$ based on the gripper orientation of $a_{\pick}$.
Our place model architecture closely follows that of Transporter Net~\cite{zeng2020transporter}. The main difference is that we explicitly encode equivariance constraints on both $\phi$ and $\psi$ networks. As a result of this change: 1) we are able to simplify the model by transposing the lifting operation $\mathcal{R}_n$ and the processing by $\phi$; 2) our new model is equivariant with respect to a larger symmetry group $C_n \times C_n$, compared to Transporter Net which is only equivariant over $C_n$. 

% Our place model architecture closely follows that of Transporter Net~\cite{zeng2020transporter}. The main differences are: 1) we explicitly encode equivariance constraints on both $\phi$ and $\psi$ networks; 2) as a result of the equivariance constraints, we are able to simplify the model by transposing the lifting operation $\mathcal{R}_n$ and the processing by $\phi$. As a result of these changes, our new model is equivariant with respect to a larger symmetry group $C_n \times C_n$, compared to Transporter Net which is only equivariant over $C_n$. 
% and 3) we obtain a place network with equivariance to a larger symmetry group $C_n \times C_n$ than Transporter Net (which is only $C_n$) reflecting symmetry of $c$ and $o_t$ independently. 

\subsubsection{Equivariant $\phi$ and $\psi$}
\label{sect:transporter_desc1}

We explicitly encode both $\phi$ and $\psi$ as $C_n$-equivariant models that satisfy the following constraints:
\begin{align}
    \label{eqn:psi}
    \psi(T^0_g(c)) &= T^0_g (\psi(c)) \\
    \label{eqn:phi}
    \phi(T^0_g(o_t)) &= T^0_g(\phi(o_t)),
    % p(a_{\place}|o_t,c) & =  \psi(\mathcal{R}_n(c)) \star \phi(o_t),
\end{align}
for $g \in \SO(2)$.
 The equivariance constraint of Equation~\ref{eqn:phi} says that when the input image rotates, we would expect the place location to rotate correspondingly. This constraint helps the model generalize across place orientations. The constraint of Equation~\ref{eqn:psi} says that when the picked object rotates (represented by the image patch $c$), then the place orientation should correspondingly rotate.

\subsubsection{Place Model}

When the equivariance constraint of Equation~\ref{eqn:psi} is satisfied, we can exchange $\mathcal{R}_n$ (the lifting operation) with $\psi$: $\psi(\mathcal{R}_n(c)) = \mathcal{R}_n(\psi(c))$. This equality is useful because it means that we only need to evaluate $\psi$ for one image patch rather than the stack of image patches $\mathcal{R}_n(c)$ -- something that is computationally cheaper. The resulting place model is then:
\begin{eqnarray}
\label{eqn:place_equiv_pre}
% f_{place}(o_t,(u,v)) = \mathcal{R}_n(\psi(c)) \star \phi(o_t).
f'_{\place}(o_t,c) & = & \mathcal{R}_n(\psi(c)) \star \phi(o_t) \\
\label{eqn:place_equiv}
& = & \Psi(c) \star \phi(o_t),
\end{eqnarray}
where Equation~\ref{eqn:place_equiv} substitutes $\Psi(c) = \mathcal{R}_n[\psi (c)]$ to simplify the expression. Here, we use $f'_{\place}$ to denote Equivariant Transporter Net defined using equivariant $\phi$ and $\psi$ in contrast to the baseline Transporter Net $f_{\place}$ of Equation~\ref{eqn:transporter1}. Note that both $f_{\place}$ and $f'_{\place}$ satisfy Proposition~\ref{prop:equivtransporter}. However, $f_{\place}$ accomplishes this by symmetrizing a non-equivariant network (i.e. evaluating $\psi(\mathcal{R}_n(c))$) whereas our model $f'_{\place}$ encodes the symmetry directly into $\psi$.

\begin{figure}[t]
    \centering
    \includegraphics[width=0.40\textwidth]{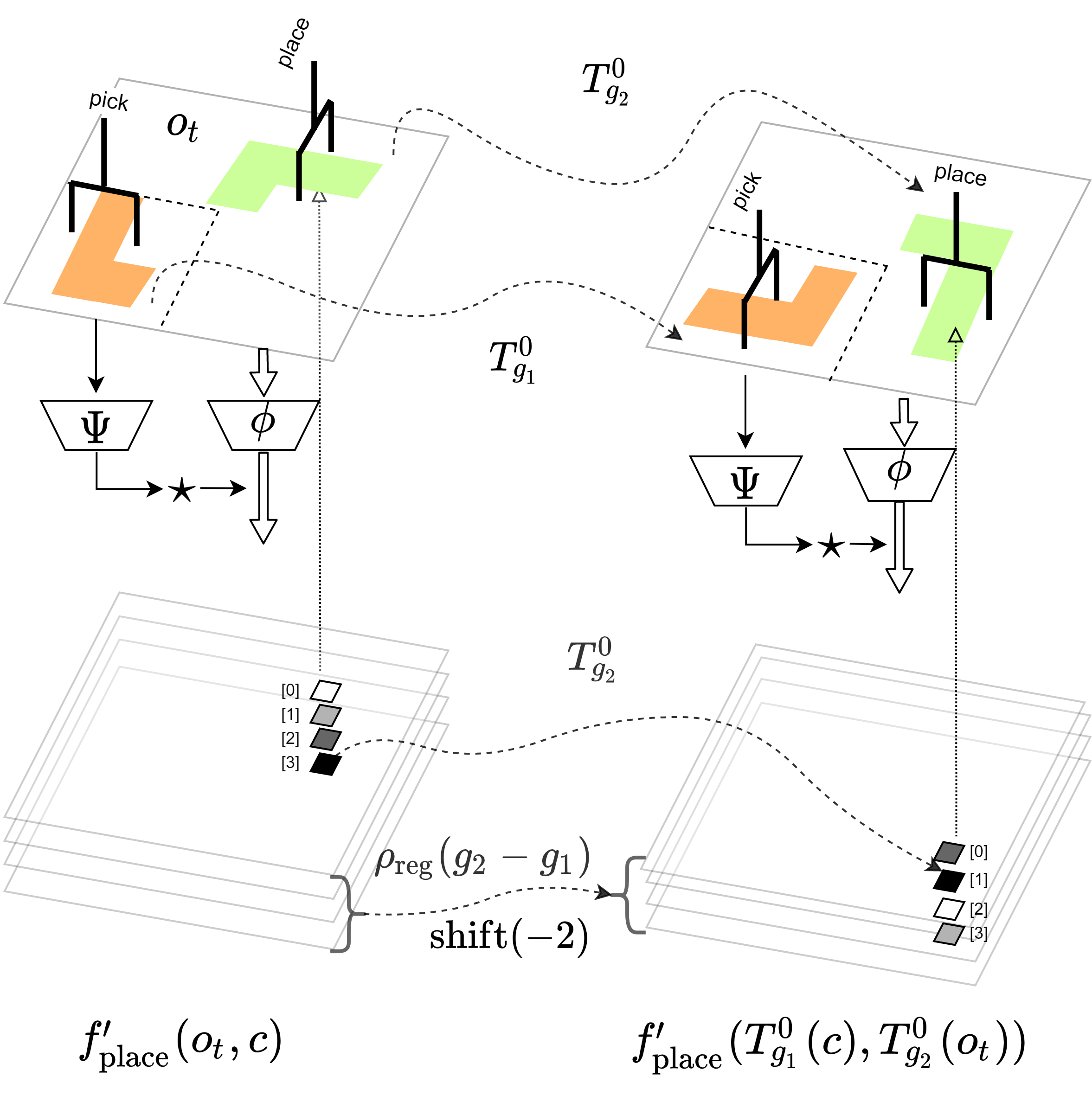}
    \caption[Place Network Architecture]{Equivariance of our placing network under the rotation of the object and the placement. A $\frac{\pi}{2}$ rotation on $c$ and a $-\frac{\pi}{2}$ rotation on $o_t\backslash c$ are equivariant to: i), a $-\frac{\pi}{2}$ rotation on the placing location, and ii), the shift on the channel of placing rotation angle from $\frac{3\pi}{2}$ (the last channel) to $\frac{\pi}{2}$ (the second channel).}
    \label{fig:place_archi}
\end{figure}

%=======================================%
%======================================%

% \subsection{Properties of Equivariant Transporter}

% Here, we summarize key properties of the proposed equivariant transporter model. Writing Equation~\ref{eqn:equi2} in terms of $f_{\place}$, we have,
% \begin{equation}
% f_{\place}(T_g^0 o_t, c) = T_g^{reg} f_{place}(o_t, c).
% \label{equi_2_pp}
% \end{equation}
% Our model also continues to satisfy Equation~\ref{eqn:transporter3} (which was satisfied by the orinal Transporter model of~\cite{zeng2020transporter}),
% \begin{equation}
% f_{\place}(o_t, T_g^0 c) = \rho_{reg}(-g) f_{place}(o_t, c).
% \label{equi_2_pp}
% \end{equation}

\subsection{Equivariance Properties of the Placing Network}

As Proposition~\ref{prop:equivtransporter} demonstrates, the baseline Transporter Net model~\cite{zeng2020transporter} encodes the symmetry that rotations of the object to be picked (represented by $c$) should result in corresponding rotations of the place orientation for that object. However, pick-conditioned place has a second symmetry that is not encoded in Transporter Net: rotations of the placement (represented by $o_t$) should also result in corresponding rotations of the place orientation. In fact, as we demonstrate in Proposition~\ref{proposition_ours} below, we encode this \emph{second type} of symmetry by enforcing the constraints of Equations~\ref{eqn:psi} and~\ref{eqn:phi}. Essentially, we go from a $C_n$-symmetric model to a $C_n \times C_n$-symmetric model.

% Although by Proposition \ref{prop:equivtransporter} the Transporter Net pick network is $C_n$-equivariant to rotations of the pick $c$, the pick and place task has greater potential symmetry in the form of simultaneous rotations of both the pick $c$ and place $o_t \setminus c$.  By enforcing equivariance in the networks $\phi$ an $\psi$, the pick network defined by Equation \ref{eqn:place_equiv} becomes $C_n \times C_n$-equivariant.

%\haojie{talk about the importance of equ 21 in manipulation tasks and can use the fancy sentence of robin's.}\\

\begin{proposition}
\label{proposition_ours}
Equivariant Transporter Net $f_{\place}'$ is $C_n \times C_n$-equivariant. That is, given rotations $g_1 \in C_n$ of the picked object and $g_2 \in C_n$ of the scene, we have that:
% Our placing network incorporates the $C_n \times C_n$ symmetry of the picking-conditioned tasks, namely, it is equivariant to both rotations of the picked objects and the rotations of the target placement:
\begin{equation}
    % \begin{aligned}
        %\psi(g \cdot Crop) \star \phi (-g \cdot O_t/Crop) =  \psi(-g \cdot Crop) \star \phi (g \cdot O_t/Crop)
        %\Psi(T^0_{g_1}(c)) \star \phi (T^0_{g_2} (o_t\backslash c)) =\\ \rho_{reg}(g_2-g_1)(T^0_{g_2}[\Psi(c) \star \phi ( o_t\backslash c)])
        f'_\place(T^0_{g_1}(c), T^0_{g_2} (o_t)) = \rho_{\reg}(g_2-g_1)T^0_{g_2} f'_\place(c, o_t).
    \label{eqn:prop2}
    % \end{aligned}
\end{equation}
% where $g_1,g_2 \in C_n$ is applied to $c$ and $o_t\backslash c$, respectively. %\eref{generalized} generalizes the equivariance of our pick-conditioned placing network, and all the property above are special case of it.
\end{proposition}
Proposition~\ref{proposition_ours} is proven in Appendix~\ref{proof_proposition} and illustrated in Figure~\ref{fig:place_archi}. The top of Figure~\ref{fig:place_archi} going left to right shows the rotation of both the object by $g_1$ (in orange) and the place pose by $g_2$ (in green). The LHS of Equation~\ref{eqn:prop2} evaluates $f'_{\place}$ for these two rotated images. The lower left of Figure~\ref{fig:place_archi} shows $f'_{\place}(c,o_t)$. Going left to right at the bottom of Figure~\ref{fig:place_archi} shows the pixel-rotation by $T^0_{g_2}$ and the channel permutation by $g_2 - g_1$ (RHS of Equation~\ref{eqn:prop2}).

% Rotation of the picked object by $g_1$ reduces the necessary rotation of the object during placement by $-g_1$, hence the shift by $\rho_\reg(-g)$ in the output.  Rotation of the target placement both increases the necessary rotation of the object by the robot by $g_2$ and moves the target placement location, hence the image rotation and angle shift of the output by $\rho_\reg(g_2) T^0_{g_2}$.

% The proof is in Appendix \ref{proof_proposition}.  The $g_1$ symmetry is proved similarly to Proposition \ref{prop:equivtransporter}. The $g_2$ symmetry depends on equivariant properties of $\phi$, $\psi$, $\mathcal{R}_n$, and $\star$.

%\textcolor{red}{Shall we mention the translation euquivariance of CNN and our model is equivariant in SE(2)?}

Note that in additional to the two rotational symmetries enforced by our model, it also has translational symmetry.  Since the rotational symmetry is realized by additional restrictions to the weights of kernels of convolutional networks, the rotational symmetry is in addition to the underlying shift equivariance of the convolutional network. Thus, the full symmetry group enforced is the group generated by $C_n \times C_n \times (\mathbb{R}^2,+)$.

Equivariant neural networks learn effectively on a lower dimensional space, the equivalence classes of samples under the group action.  Thus a larger group results in an even smaller dimensional sample space and thus better coverage by the training data.

%\vspace{-1cm}
\begin{figure*}[ht]
\vspace{-1cm}
    \centering
    \includegraphics[width=1\textwidth]{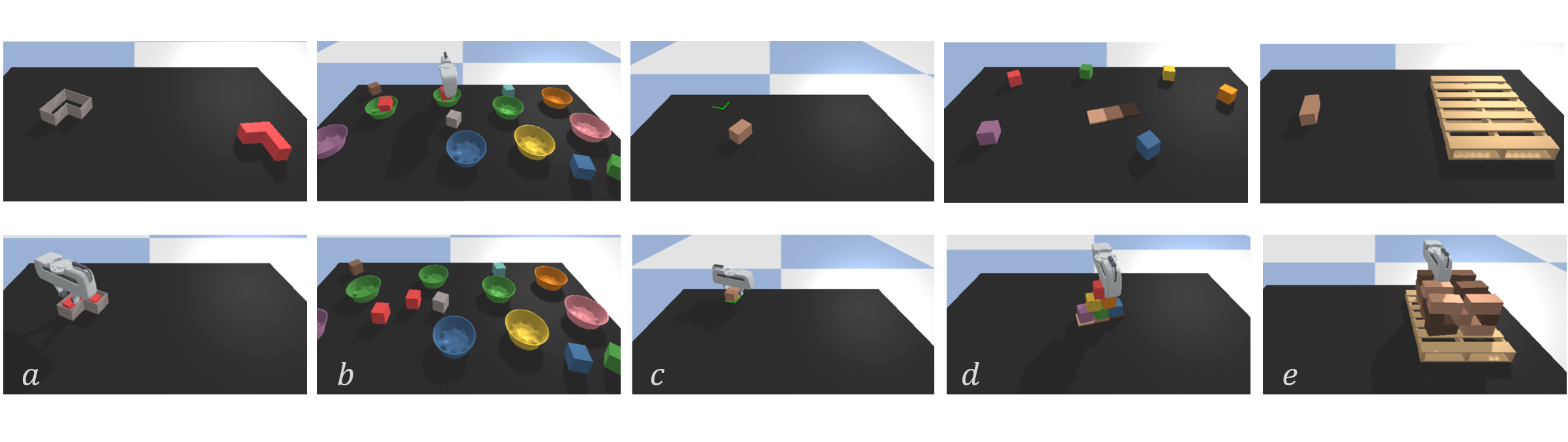}
    \caption{ Simulated environment for parallel-jaw gripper tasks. From left to right: (a) inserting blocks into fixtures, (b) placing red boxes into green bowls, (c) align box corners to green lines, (d) stacking a pyramid of blocks, (e) palletizing boxes.}
    \label{fig:gripper_env}
\end{figure*}
\subsection{Model Architecture Details}

\subsubsection{Pick model $f_p$ (Equation~\ref{eqn:pick1})}

The input to $f_p$ is a $4$-channel RGB-D image $o_t \in \mathbb{R}^{4\times H \times W}$. The output is a feature map $p(u,v) \in \mathbb{R}^{H\times W}$ which encodes a distribution over pick location. $f_p$ is implemented as an 18-layer euqivariant residual network with a U-Net~\cite{ronneberger2015u} as the main block. The U-net has 8 residual blocks (each block contains 2 equivariant convolution layers~\cite{weiler2019general} and one skip connection): 4 residual blocks~\cite{he2016deep} are used for the encoder and the other 4 residual blocks are used for the decoder. The encoding process trades spatial dimensions for channels with max-pooling in each block; the decoding process upsamples the feature embedding with bilinear-upsampling operations. The first layer maps the trivial representation of $o_t$ to regular representation and the last equivariant layer transforms the regular representation back to the trivial representation, followed by image-wide softmax. ReLU activations~\cite{nair2010rectified} are interleaved inside the network. 

\subsubsection{Pick model $f_\theta$ (Equation~\ref{eqn:pick2})}

Given the picking location $(u^*,v^*)$, the pick angle network $f_\theta$ takes as input a crop $c\in \mathbb{R}^{4\times H_1 \times W_1}$ centered on $(u^*,v^*)$ and outputs the distribution $p(\theta|u,v) \in \mathbb{R}^{n/2}$, where $n$ is the size of the rotation group(i.e. $n=| C_n |$). The first layer maps the trivial representation of $c$ to a quotient regular representation followed by 3 residual blocks containing max-pooling operators. This goes to two equivariant convolution layers and then to an average pooling layer.

\subsubsection{Place models $\phi$ and $\psi$}

Our place model has two equivariant convolution networks, $\phi$ and $\psi$, and both have similar architectures to $f_p$. The network $\phi$ takes as input a zero-padded version of $o_t$, $\mathrm{pad}(o_t)\in \mathbb{R}^{4\times (H+d) \times (W+d)}$, and generates a dense feature map, $\phi(\mathrm{pad}(o_t))\in\mathbb{R}^{(H+d) \times (W+d)}$, where $d$ is the padding size. The network $\psi$ takes as input the image patch $c \in \mathbb{R}^{4\times H_2 \times W_2}$ and outputs $\psi(c)\in \mathbb{R}^{H_2 \times W_2}$. After applying rotations of $C_n$ to $\psi(c)$, the transformed dense embeddings $\Psi(c)\in \mathbb{R}^{n\times H_2 \times W_2}$ are cross-correlated with $\phi(\mathrm{pad}(o_t))$ to generate the placing action distribution $p(a_{\place}|o_t,a_{\pick}) \in \mathbb{R}^{n\times H \times W}$, where the channel axis $n$ corresponds to placing angles, $\frac{2\pi i}{n}$ for $0\leq i < n$.

\subsubsection{Group Types and Sizes}

The networks $f_p$, $\psi$, and $\phi$ are all defined using $C_6$ regular representations. The network $f_\theta$ is defined using the regular quotient representation $C_{36}/C_2$, which corresponds to the number of allowed place orientations.

% We select rotation group $C_6$ for our picking location network, $\psi$ network and $\phi$ network, and $C_{36}$ for our picking angle model and rotations on $\psi(c)$. 

\subsubsection{Training Details:}
\label{traing_settings}

We train Equivariant Transporter Network with the Adam~\cite{kingma2014adam} optimizer with a fixed learning rate of $10^{-4}$. It takes about 0.8 seconds to complete one SGD step with a batch size of 1 on a NVIDIA Tesla V100 SXM2 GPU. For each task, we evaluate performance every 10k steps on 100 unseen tests. On most tasks, the best performance is achieved in less than 10k SGD steps. Our model converges in a few hours on all tasks.

\section{Experiments}

We evaluate Equivariant Transporter using the Ravens-10 Benchmark~\cite{zeng2020transporter} and variations thereof.

% We evaluate Equivariant Transporter using the Ravens-10 Benchmark~\cite{zeng2020transporter} that includes a collection of simulated tasks in Pybullet~\cite{coumans2016pybullet} with a suction gripper. To evaluate our model in the parallel-jaw gripper setting, we update five of the ten Ravens-10 tasks to incorporate the gripper. We also demonstrate our method on a physical robot with three Ravens-10 tasks.

% To verify our model's performance with a parallel-jaw gripper that requires learning the pick-angle, we select 5 tasks from the Raven-10 Benchmark and provide a simulated learning environment that is feasible to two-finger gripper manipulation. We also validate our method with a real robot to demonstrate its practical value. 

\subsection{Tasks}

\subsubsection{Ravens-10 Tasks}

Ravens-10 is a behaviour cloning simulation environment for manipulation, where each task owns an oracle that can sample expert demonstrations from the distribution of successful picking and placing actions with the access to the ground-truth pose of each object. The 10 tasks of Ravens can be classified into 3 categories: \textit{Single-object manipulation tasks} (block-insertion, align-box-corner); \textit{Multiple-object manipulation tasks} (place-red-in-green, towers-of-hanoi, stack-block-pyramid, palletizing-boxes, assembling-kits, packing-boxes); \textit{Deformable-object manipulation task} (manipulating-rope, sweeping-piles). Detailed explanations of the tasks can be found in Appendix~\ref{raven_10_details}

% \subsection{Simulated Environment for Parallel-jaw Gripper}
\subsubsection{Ravens-10 Tasks Modified for the Parallel Jaw Gripper}
\label{parallel env}

We selected 5 tasks (block-insertion, align-box-corner, place-red-in-green, tack-block-pyramid, palletizing-boxes) from Ravens-10 and replaced the suction cup with the Franka Emika gripper. \figref{fig:gripper_env} illustrates the initial state and completion state for each of these five tasks. For each of these five tasks, we defined an oracle agent. Since the Transporter Net framework assumes that the object does not move during picking, we defined these expert generators such that this was the case.

% To avoid the movement of the target object caused by grasping with the parallel-jaw gripper, we shrink the successful picking location distribution to the center line of symmetrical objects. The expert planner defined on continuous picking angle space can generate successful immobile picking actions but it is unavoidable for the planner defined on discretized picking angle space (e.g. $C_{36}$). To tackle it, we only collect successful demonstrations generated by discretized planners, i.e., demonstrations with tiny movements during grasping.
%%%%%%%%%%%%%%%%%%%%%%%%%%%%%%%%%%%%%%
%---------------table1--------------%
\begin{table*}[h]
  \setlength\tabcolsep{4pt}
  \begin{center}
  \scriptsize
  \begin{tabular}{@{}lcccccccccccccccccccc@{}}
  \toprule
  & \multicolumn{4}{c}{block-insertion} & \multicolumn{4}{c}{place-red-in-green} & \multicolumn{4}{c}{towers-of-hanoi} & \multicolumn{4}{c}{align-box-corner} & \multicolumn{4}{c}{stack-block-pyramid} \\
  \cmidrule(lr){2-5} \cmidrule(lr){6-9} \cmidrule(lr){10-13} \cmidrule(lr){14-17} \cmidrule(lr){18-21}
  Method & 1 & 10 & 100 & 1000 & 1 & 10 & 100 & 1000 & 1 & 10 & 100 & 1000 & 1 & 10 & 100 & 1000 & 1 & 10 & 100 & 1000 \\
  \midrule
  Equivariant Transporter                & \textbf{100} & \textbf{100} & \textbf{100} & \textbf{100} & \textbf{98.5} & \textbf{100} & \textbf{100} & \textbf{100} & \textbf{88.1} & \textbf{95.7} & \textbf{100} & \textbf{100} & 41.0 & \textbf{99.0} & \textbf{100} & \textbf{100} & \textbf{34.6} & \textbf{80.0} & \textbf{90.8} & \textbf{95.1} \\
  Transporter Network                  & \textbf{100} & \textbf{100} & \textbf{100} & \textbf{100} &  84.5 & \textbf{100} & \textbf{100} & \textbf{100} & 73.1 & 83.9 & 97.3 & 98.1 & 35.0 & 85.0 & 97.0 & 98.0 & 13.3 & 42.6 & 56.2 & 78.2 \\
  Form2Fit       & 17.0 & 19.0 & 23.0 & 29.0 & 83.4 & \textbf{100} & \textbf{100} & \textbf{100} & 3.6 & 4.4 & 3.7 & 7.0 & 7.0 & 2.0 & 5.0 & 16.0 & 19.7 & 17.5 & 18.5 & 32.5 \\
  Conv. MLP                           & 0.0 & 5.0 & 6.0 & 8.0 & 0.0 & 3.0 & 25.5 & 31.3 & 0.0 & 1.0 & 1.9 & 2.1 & 0.0 & 2.0 & 1.0 & 1.0 & 0.0 & 1.8 & 1.7 & 1.7 \\
  GT-State MLP                        & 4.0 & 52.0 & 96.0 & 99.0 & 0.0 & 0.0 & 3.0 & 82.2 & 10.7 & 10.7 & 6.1 & 5.3 & 47.0 & 29.0 & 29.0 & 59.0 & 0.0 & 0.2 & 1.3 & 15.3 \\
  GT-State MLP 2-Step                        & 6.0 & 38.0 & 95.0 & \textbf{100} & 0.0 & 0.0 & 19.0 & 92.8 & 22.0 & 6.4 & 5.6 & 3.1 & \textbf{49.0} & 12.0 & 43.0 & 55.0 & 0.0 & 0.8 & 12.2 & 17.5 \\
  \midrule
  & \multicolumn{4}{c}{palletizing-boxes} & \multicolumn{4}{c}{assembling-kits} & \multicolumn{4}{c}{packing-boxes} & \multicolumn{4}{c}{manipulating-rope} & \multicolumn{4}{c}{sweeping-piles}\\
  \cmidrule(lr){2-5} \cmidrule(lr){6-9} \cmidrule(lr){10-13} \cmidrule(lr){14-17} \cmidrule(lr){18-21}
  & 1 & 10 & 100 & 1000 & 1 & 10 & 100 & 1000 & 1 & 10 & 100 & 1000 & 1 & 10 & 100 & 1000 & 1 & 10 & 100 & 1000 \\
  \midrule
  Equivariant Transporter                & \textbf{75.3} & \textbf{98.9} & \textbf{{99.6}} & \textbf{{99.6}} & \textbf{63.8} & \textbf{90.6} & \textbf{{98.6}} & \textbf{{100}} & \textbf{{98.3}} & \textbf{{99.4}} & \textbf{{99.6}} & \textbf{{100}} & \textbf{31.0} & \textbf{85.0} & \textbf{92.3} & \textbf{98.4} & \textbf{97.9} & \textbf{99.5} & \textbf{100} & \textbf{100} \\
  Transporter Network                 & 63.2 & 77.4 & 91.7 & 97.9 & 28.4 & 78.6& 90.4& 94.6 & 56.8 & 58.3 & 72.1 & 81.3 & 21.9 & 73.2 & 85.4 & 92.1 & 52.4 & 74.4 & 71.5 & 96.1 \\
  Form2Fit                            & 21.6 & 42.0 & 52.1 & 65.3 & 3.4  & 7.6 & 24.2& 37.6 & 29.9 & 52.5 & 62.3 & 66.8 & 11.9 & 38.8 & 36.7 & 47.7 & 13.2 & 15.6 & 26.7 & 38.4 \\
  Conv. MLP                           & 31.4 & 37.4 & 34.6 & 32.0 & 0.0  & 0.2 & 0.2 & 0.0 & 0.3 & 9.5 & 12.6 & 16.1 & 3.7 & 6.6 & 3.8 & 10.8 & 28.2 & 48.4 & 44.9 & 45.1 \\
  GT-State MLP                        & 0.6  & 6.4  & 30.2 & 30.1 & 0.0  & 0.0 & 1.2 & 11.8 & 7.1 & 1.4 & 33.6 & 56.0 & 5.5 & 11.5 & 43.6 & 47.4 & 7.2 & 20.6 & 63.2 & 74.4 \\
  GT-State MLP 2-Step                 & 0.6  & 9.6  & 32.8 & 37.5 & 0.0  & 0.0 & 1.6 & 4.4 & 4.0 & 3.5 & 43.4 & 57.1 & 6.0 & 8.2 & 41.5 & 58.7 & 9.7 & 21.4 & 66.2 & 73.9 \\
  \bottomrule
  \end{tabular}
  \end{center}
  \vspace{0.5em}
  \caption{\scriptsize\textbf{Performance comparisons on Ravens-10 benchmark (suction gripper).} Success rate (mean\%) vs. the number of demonstration episodes (1, 10, 100, or 1000) used in training. Best performances are highlighted in bold.}
  \vspace{-1.0em}
  \label{table:sample-efficiency1}
\end{table*}
%%%%%%%%%%%%%%%%%%%%%%%%%%%%%%%%%%%%%%

%%%%%%%%%%%%%%%%%%%%%%%%%%%%%%%%%%%%%%
%---------------table1--------------%
\begin{table*}[h]
  \setlength\tabcolsep{4pt}
  \centering
  \scriptsize
  \begin{tabular}{@{}lccccccccccccccc@{}}
  \toprule
  & \multicolumn{3}{c}{block-insertion} & \multicolumn{3}{c}{place-red-in-green} & \multicolumn{3}{c}{palletizing-boxes} & \multicolumn{3}{c}{align-box-corner} & \multicolumn{3}{c}{stack-block-pyramid} \\
  \cmidrule(lr){2-4} \cmidrule(lr){5-7} \cmidrule(lr){8-10} \cmidrule(lr){11-13} \cmidrule(lr){14-16}
  Method & {1} & 10 & 100 &  1 & 10 & 100  & 1 & 10 & 100  & 1 & 10 & 100  & 1 & 10 & 100\\
  \midrule
  Equivariant Transporter                & \textbf{100} & \textbf{100} & \textbf{100} & \textbf{95.6} & \textbf{100} & \textbf{100} & \textbf{96.1} & \textbf{100} & \textbf{100} & \textbf{64.0} & \textbf{99.0} & \textbf{100} & \textbf{62.1} & \textbf{85.6} & \textbf{98.3} \\
  
  Transporter Network                  & 98.0 & \textbf{100} & \textbf{100}  & 82.3 & 94.8 & \textbf{100} &  84.2 & 99.6 & \textbf{100} & 45.0 & 85.0 & 99.0 & 16.6 & 63.3 & 75.0  \\
  
  \bottomrule
  \end{tabular}
  \vspace{0.5em}
  \caption{\scriptsize\textbf{Performance comparisons on tasks with a parallel-jaw end effector.} Success rate (mean\%) vs. the number of demonstration episodes (1, 10, or 100) used in training.}
  \vspace{-1.0em}
  \label{table:sample-efficiency2}
\end{table*}
%%%%%%%%%%%%%%%%%%%%%%%%%%%%%%%%%%%%%%

\subsection{Training and Evaluation}

\subsubsection{Training}

For each task, we produce a dataset of $n$ expert demonstrations, where each demonstration contains a sequence of one or more observation-action pairs $(o_t,\Bar{a}_t)$. Each action $\Bar{a}_t$ contains an expert picking action $\Bar{a}_{\pick}$ and an expert placing action $\Bar{a}_{\place}$. We use $\Bar{a}_t$ to generate one-hot pixel maps as the ground-truth labels for our picking model and placing model. The models are trained using a cross-entropy loss.

\subsubsection{Metrics}

We measure performance the same way as it was measured in ~\cite{zeng2020transporter} -- using a metric in the range of 0 (failure) to 100 (success). Partial scores are assigned to multiple-action tasks. For example, in the block-stacking task where the agent needs to construct a 6-block pyramid, each successful rearrangement is credited with a score of 16.667. We report highest validation performance during training, averaged over 100 unseen tests for each task. 

\subsubsection{Baselines} 

We compare our method against Transporter Net~\cite{zeng2020transporter} as well as the following baselines previously used in the Transporter Net paper~\cite{zeng2020transporter}. \textit{Form2Fit}~\cite{zakka2020form2fit} introduces a matching module with the measurement of $L_2$ distance of high-dimension descriptors of picking and placing locations. \textit{Conv-MLP} is a common end-to-end model~\cite{levine2016end} which outputs $a_{\pick}$ and  $a_{\place}$ using convolution layers and MLPs (multi-layer perceptrons). \textit{GT-State MLP} is a regression model composed of an MLP that accepts the ground-truth poses and 3D bounding boxes of objects in the environment. \textit{GT-State MLP 2-step} outputs the actions sequentially with two MLP networks and feeds $a_{\pick}$ to the second step. All regression baselines learn mixture densities~\cite{bishop1994mixture} with log likelihood loss.

\subsubsection{Adaptation of Transporter Net for Picking Using a Parallel Jaw Gripper}

In order to compare our method against Transporter Net for the five parallel jaw gripper tasks, we must modify Transporter to handle the gripper. We accomplish this by~\cite{zeng2018learning} lifting the input scene image over $C_n$, producing a stack of differently oriented input images which is provided as input to the pick network $f_{\pick}$. The results are counter-rotated at the output of $f_{\pick}$.

% To fit Transporter Network to parallel-jaw gripper tasks, the input observation is first rotated n times by $C_n$ and the mini-batch of $n$ images is fed to the picking network to output $p(a_{\pick})\in\mathbb{R}^{n\times H\times W}$. Then, each channel of the output undergoes a reversed rotation by $-C_n$.

\subsection{Results for the Ravens-10 Benchmark Tasks}

\subsubsection{Task Success Rates} 
%=======================================%
%======================================%
\begin{figure}
    \centering
     \subfigure[]{ \includegraphics[clip,width=0.23\textwidth]{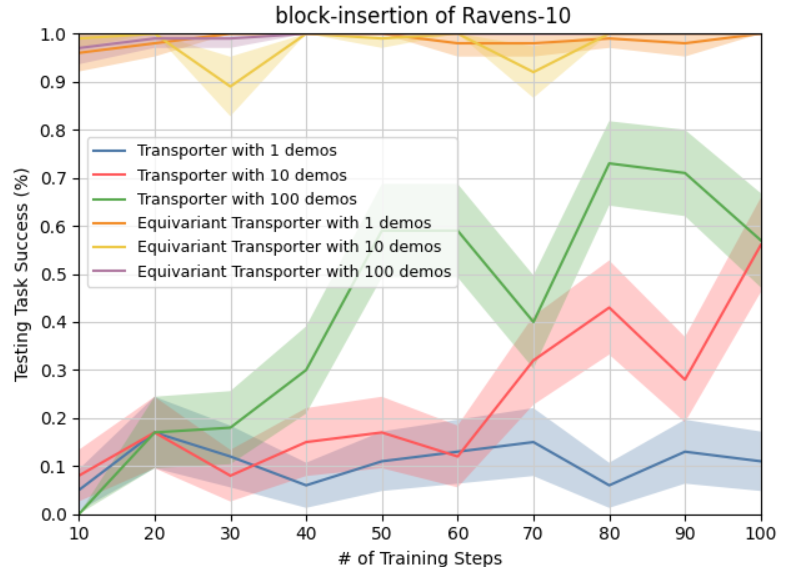}}
     %\hspace{0.1cm}
     \subfigure[]{ \includegraphics[clip,width=0.232\textwidth]{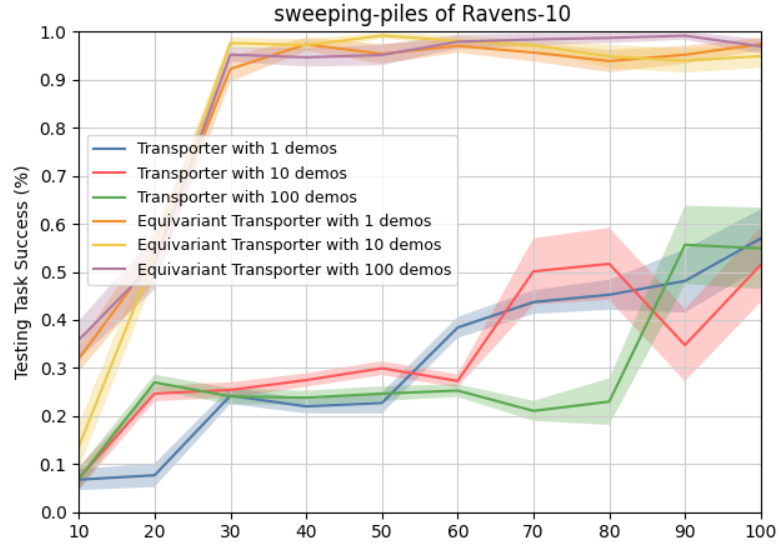}}
    \caption[faster learning]{Equivariant Transporter Network converges faster than Transporter Network. Left: Block-insertion task. Right: sweeping-piles task. On the block insertion task, Equivariant Transporter can hit greater than $90\%$ success rate after 10 training steps and achieve $100\%$ succeess rate with less than 100 training steps.}
    \label{fig:fast_converge}
\end{figure}

Table~\ref{table:sample-efficiency1} shows the performance of our model on the Raven-10 tasks for different numbers of demonstrations used during training. All tests are evaluated on unseen configurations, i.e., random poses of objects, and three tasks (align-box-corner, assembling-kits, packing-box) use unseen objects. 
% Although a small number of stochastic expert demonstrations makes this benchmark challenging,  only with partial equivariance, Transporter Network can achieve better sample efficiency than other baselines.
% Having a generalized equivariance, Equivariant Transporter leverages more desirable symmetries of rearrangement manipulation tasks, and thus outperforms Transporter Network by orders of magnitude more sample efficiency. 
Our proposed Equivariant Transporter Net outperforms all the other baselines in nearly all cases. The amount by which our method outperforms others is largest when the number of demonstrations is smallest, i.e. with only 1 or 10 demonstrations. With just 10 demonstrations per task, our method can achieve $\geq 95\%$ success rate on 7/10 tasks. 

\subsubsection{Training Efficiency}

Another interesting consequence of our more structured model is that training is much faster. Figure~\ref{fig:fast_converge} shows task success rates as a function of the number or SGD steps for two tasks (Block Insertion and Sweeping Piles). Our equivariant model converges much faster in both cases.

% The gap in performance between our model and Transporter Network occurs when 10 or 100 demonstrations are used in training. Our model can achieve greater than 95\% success rate on 7/10 tasks after being trained with 10 demonstrations while Transporter Network can only achieve the same goal on 2/10 tasks. Although Transporter Network can achieve 100\% success rate on block insertion task, we show that in \figref{fig:fast_converge} Equivariant Transporter converges with a smaller number of training step than Transporter Network. Besides, our model hits 100\% success rate in most tasks with less than 1000 training demonstrations. 

\subsection{Results for Parallel Jaw Gripper Tasks}

\subsubsection{Task Success Rates}

Table~\ref{table:sample-efficiency2} compares the performance of Equivariant Transporter with the baseline Transporter Net for the Parallel Jaw Gripper tasks. Again, our method outperforms the baseline in nearly all cases.

\subsubsection{Comparison with Ravens-10}

One interesting observation that can be made by comparing Tables~\ref{table:sample-efficiency1} and~\ref{table:sample-efficiency2} is that both Equivariant Transporter and baseline Transporter do better on the gripper versions of the task compared to the original Ravens-10 versions. This is likely caused by the fact that the expert demonstrations we developed for the gripper version on the task have less stochastic gripper poses during pick than the case in the original Ravens-10 benchmark.

% The performances on palletizing-boxes task and stacking-block-pyramid are higher than these in Table~\ref{table:sample-efficiency1}. We believe this is due to two reasons: i) the expert's picking location distribution is constrained to the center lines of the objects and the picking demonstrations are less stochastic than before. It will further cause the distribution of picking-conditioned placing to be less stochastic. As shown in previous work~\cite{zeng2020transporter,machado2018revisiting}, a deterministic policy is easier to learn than a stochastic one. ii) parallel-jaw gripper has 180\degree symmetry and fits center-line grasps. Both Table~\ref{table:sample-efficiency1} and Table~\ref{table:sample-efficiency2} indicate the advantage of our proposed model that is more sample efficiency and can generalize better than a variety of baselines. 

\subsection{Ablation Study}
\label{ablation_study}

\subsubsection{Ablations} 

We performed an ablation study to evaluate the relative importance of the equivariant models in pick ($f_p$ and $f_\theta$) and place ($\psi$ and $\phi$). We compare four versions of model with various levels of equivariance: non-equivariant pick and non-equivariant place (baseline Transporter), equivariant pick and non-equivariant place, 
non-equivariant pick and equivariant place, and
equivariant pick and equivariant place (Equivariant Transporter).

% We conduct ablation study to evaluate the performance of our $C_n \times C_n$-equivariant placing network by introducing two hybrid models:
% \begin{enumerate}
%     \item \textbf{Equivaiant Transporter Pick + Transporter Place.} This model is composed of the pick network of Equivariant Transporter Net and the place network of Transporter Net.
%     \item \textbf{Transporter Pick + Equivaiant Transporter Place.} This model is composed of the pick network of Transporter Net and the place network of ours.
% \end{enumerate}
\begin{figure}[t]
    \centering
     \subfigure[]{ \includegraphics[width=0.23\textwidth]{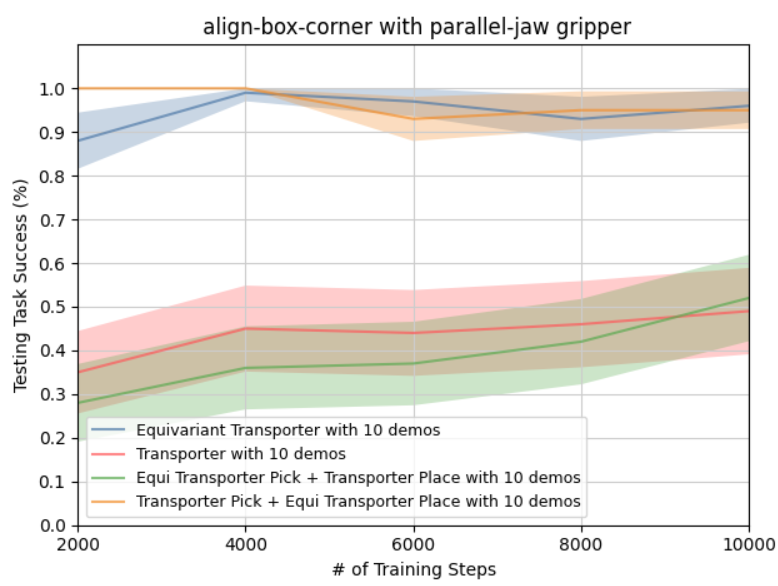}}
     %\hspace{0.1cm}
     \subfigure[]{ \includegraphics[width=0.232\textwidth]{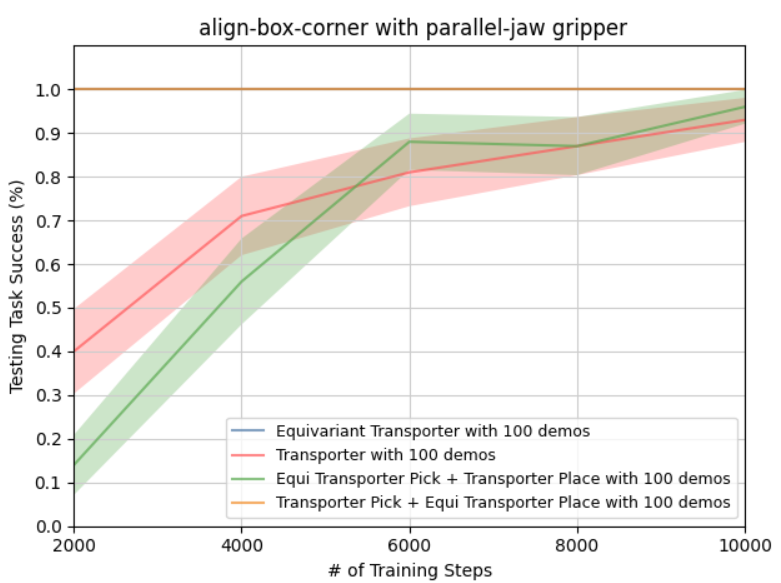}}
    \vfill
    \subfigure[]{ \includegraphics[width=0.23\textwidth]{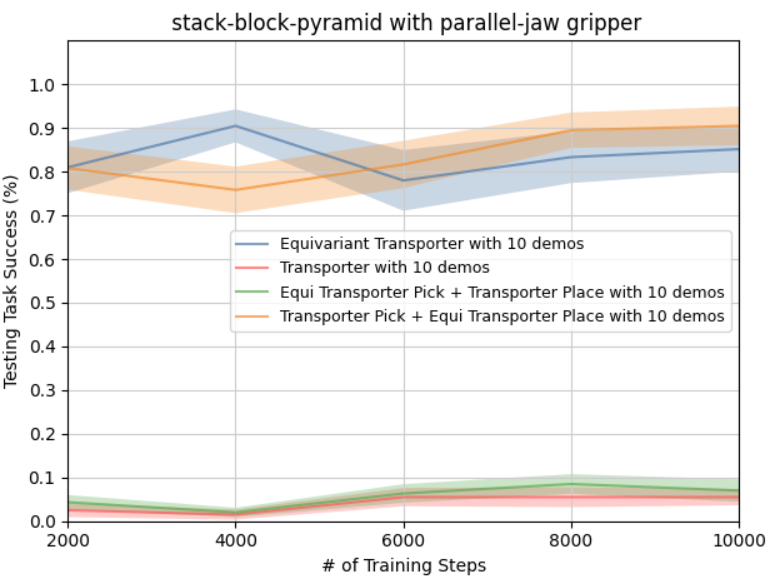}}
     %\hspace{0.1cm}
     \subfigure[]{ \includegraphics[width=0.232\textwidth]{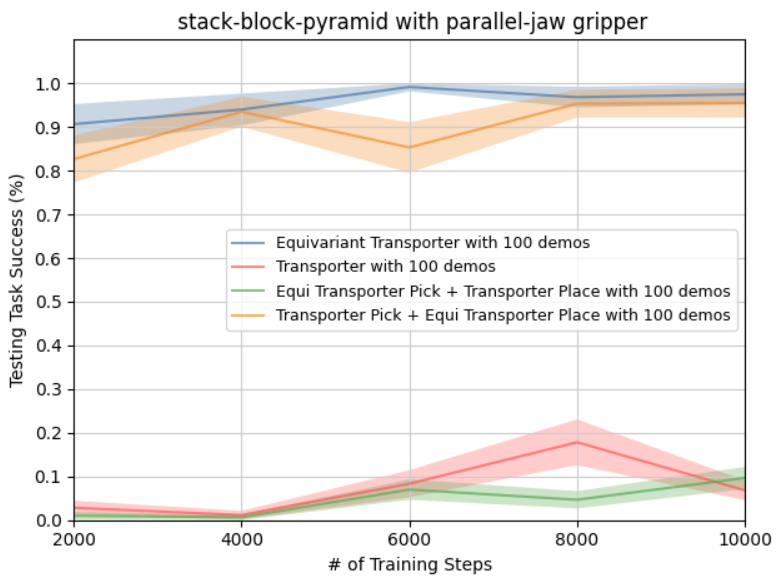}}
    \caption[Ablation Study]{Ablation study. Performance is evaluated on 100 unseen tests of each task.}
    \label{fig:ablation}
\end{figure}

\subsubsection{Results} 

\figref{fig:ablation} shows the performance of all four ablations as a function of the number of SGD steps for the scenario where the agent is given 10 or 100 expert demonstrations. The results indicate that place equivariance (i.e. equivariance of $\psi$ and $\phi$) is namely responsible for the gains in performance of Equivariant Transporter versus baseline Transporter. This finding is consistent with the argument that it is the larger $C_n \times C_n$ symmetry group (only present with equivariant place) that is responsible for our performance gains. Though the non-equivariant and equivariant pick networks result in comparable performance, the equivariant network is far more computationally efficient, taking a single image as input versus 36 for the non-equivariant network.

%=======================================%
%======================================%
\begin{figure}[]
    \centering
    \centering
     \subfigure[]{ \includegraphics[clip,width=0.23\textwidth]{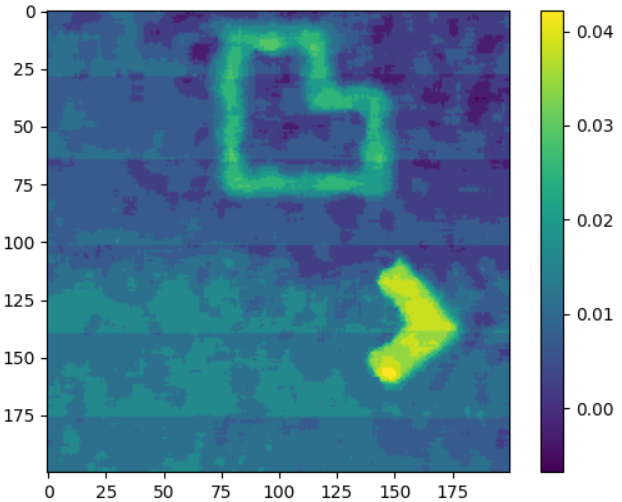}}
     %\hspace{0.1cm}
     \subfigure[]{ \includegraphics[clip,width=0.226\textwidth]{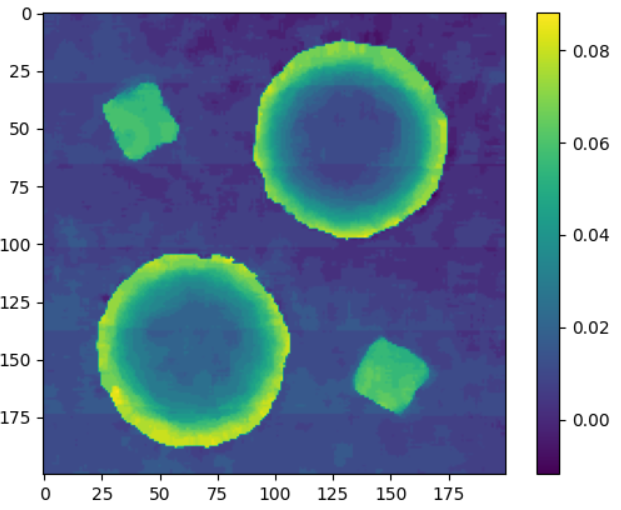}}
    \caption[depth image]{Real robot experiment: initial observation $o_t\in\mathbb{R}^{200\times200}$ from the depth sensor. The left figure shows the block insertion task; the right figure shows the task of placing boxes in bowls. The depth value ($\mathrm{meter}$) is illustrated in the color bar.}
    \label{fig:depth}
    \vspace{-0.5cm}
\end{figure}

\begin{figure}
    \centering
    \includegraphics[width=0.48\textwidth]{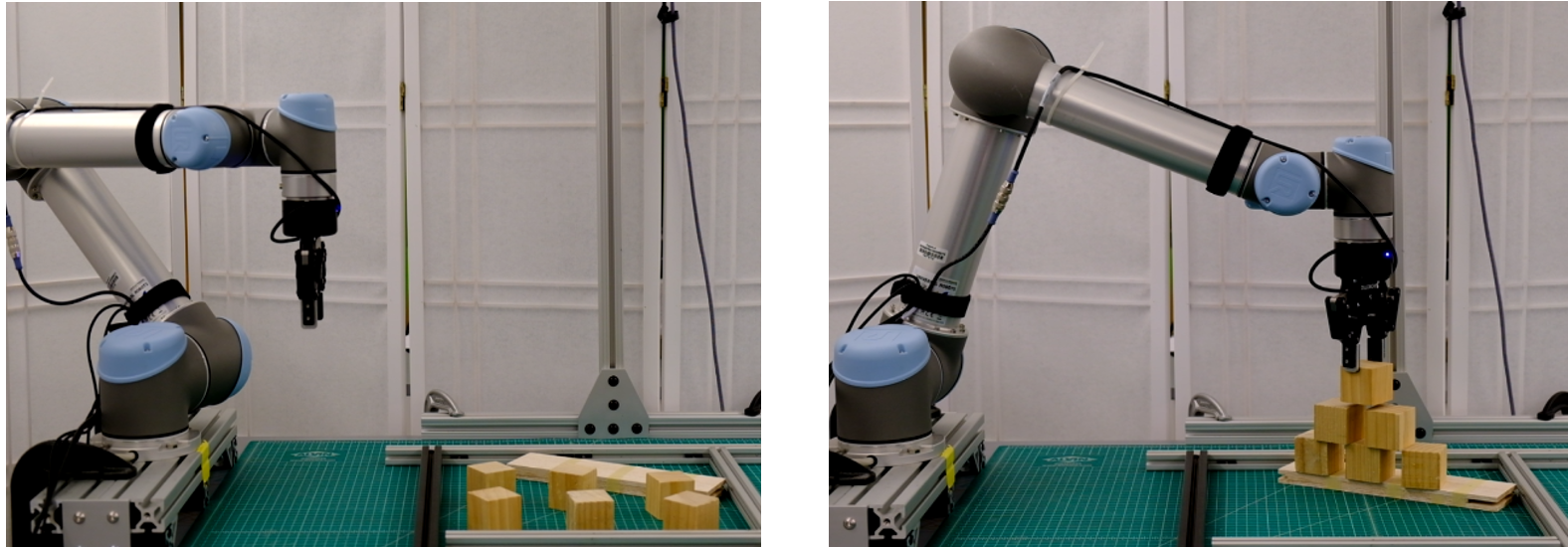}
    \caption[depth image]{Stack-block-pyramid task on the real robot. The left figure shows the initial state; the right figure shows the completion state.}
    \label{fig:robot}
\end{figure}

\subsection{Experiments on a Physical Robot}

We evaluated Equivariant Transporter on a physical robot in our lab. There was no use of the simulator in this experiment -- all demonstrations were performed on the real robot.

\subsubsection{Setup} 

We used a UR5 robot with a Robotiq-85 end effector. The workspace was a $40cm \times 40cm$ region on a table beneath the robot (see Figure~\ref{fig:robot}). The observations $o$ were $200 \times 200$ depth images obtained using a Occipital Structure Sensor that was mounted pointing directly down at the table (see Figure~\ref{fig:depth}). 

\subsubsection{Tasks}

We evaluated Equivariant Transporter on three of the Ravens-10 gripper-modified tasks: block insertion, placing boxes in bowls, and stacking a pyramid of blocks. Since our sensor only measures depth (and not RGB), we modified the box-in-bowls task such that box color was irrelevant to success, i.e. the task is simply to put any box into a bowl.

\subsubsection{Demonstrations}

We obtained 10 human demonstrations of each task. These demonstrations were obtained by releasing the UR5 brakes and pushing the arm physically so that the harmonic actuators were back-driven.

\begin{table}[H]
    \centering
    \begin{tabular}{c|c|c|c}
    \hline
         Task & \# demos & \# completions / \# trials  & success rate\\
        \hline
         stack-block-pyramid & 10 & 17/20  & 95.8\%\\
        \hline
        place-box-in-bowl & 10 & 20/20  & 100\%\\
        \hline
        block-insertion & 10 & 20/20  & 100\%\\
     \hline
    \end{tabular}
    \caption{Task success rates for physical robot evaluation tasks.}
    \label{tab:real_test_performance}
\end{table}

\subsection{Training and Evaluation}

For each task, our model was trained for 10k SGD steps. During testing, objects were randomly placed on the table. A task was considered to have failed when a single incorrect pick or place occurred. We evaluated on 20 unseen configurations of each task.

\subsubsection{Results}

Table\:\ref{tab:real_test_performance} shows results from 20 runs of each of the three tasks. Notice that the success rates here are higher than they were for the corresponding tasks performed in simulation (Table~\ref{table:sample-efficiency2}). This is likely caused by the fact that the criteria for task success in simulation (less than 1 cm translational error and less than $\frac{\pi}{12}$ rotation error were more conservative than is actually the case in the real world. Videos can be found in supplementary materials.

% Since we cannot penalize $1$cm translation errors and $\frac{\pi}{12}$ rotation errors relative to target poses similar to simulation, the success rate of stacking task is higher than simulation tests. 

\section{Conclusion and Limitations}
\label{sec:conclusion}

This paper explores the symmetries present in the pick and place problem and finds that they can be described by the direct product group $C_n \times C_n$, where $C_n$ denotes the cyclic group of discrete orientations. This corresponds to the group of different pick and place orientations. We evaluate the Transporter Network model proposed in~\cite{zeng2020transporter} and find that it encodes one of these symmetries (the pick symmetry), but not the other (the place symmetry). We propose a novel version of Transporter Net, Equivariant Transporter Net, which we show encodes both types of symmetries. We evaluate our model on the Ravens-10 Benchmark and evaluate against multiple strong baselines. Finally, we demonstrate that the method can effectively be used to learn manipulation policies on a physical robot. One limitation of our framework as it is presented in this paper is that it relies entirely on behavior cloning. A clear direction for future work is to integrate more on-policy learning which we believe would enable us to handle more complex tasks.

\section*{Acknowledgement}
This work was supported in part by NSF 1724257, NSF 1724191, NSF
1763878, NSF 1750649, NSF 2107256, NSF 2134178, NASA 80NSSC19K1474, the Harold Alfond Foundation, and the Roux Institute.

% compare Equivariant Transporter against a number of baselines and find that it 

% and proposes a novel Equivariant version of Transporter Network~\cite{zeng2020transporter}. We find that the the model proposed in~\cite{zeng2020transporter} encodes symmetries over pick orientation but not symmetries over place. 

% -- this paper proposes a novel version of Transporter Net~\cite{zeng2020transporter} that incorporates equivariant layers into the $\phi$ and $\psi$ models used to infer place location. 

% In this work, we explored the desirable symmetries in 2D manipulation rearrangement tasks and extend the partial equivariance inside Transporter Network to a more generalized equivariant network. Our proposed model, Equivariant Transporter Network, highly improves the sample efficiency in various tasks and settings. Two limitations of our method could be further research directions: i) actions are defined on descretized SE(2) space instead of continuous space; ii) SE(3)-Equivariance of rearrangement manipulation tasks is not investigated in our work.

%\section*{Acknowledgments}

%% Use plainnat to work nicely with natbib. 

\bibliographystyle{plainnat}
\bibliography{references}

\begin{thebibliography}{35}
\providecommand{\natexlab}[1]{#1}
\providecommand{\url}[1]{\texttt{#1}}
\expandafter\ifx\csname urlstyle\endcsname\relax
  \providecommand{\doi}[1]{doi: #1}\else
  \providecommand{\doi}{doi: \begingroup \urlstyle{rm}\Url}\fi

\bibitem[Berscheid et~al.(2020)Berscheid, Mei{\ss}ner, and
  Kr{\"o}ger]{berscheid2020self}
Lars Berscheid, Pascal Mei{\ss}ner, and Torsten Kr{\"o}ger.
\newblock Self-supervised learning for precise pick-and-place without object
  model.
\newblock \emph{IEEE Robotics and Automation Letters}, 5\penalty0 (3):\penalty0
  4828--4835, 2020.

\bibitem[Besl and McKay(1992)]{besl1992method}
Paul~J Besl and Neil~D McKay.
\newblock Method for registration of 3-d shapes.
\newblock In \emph{Sensor fusion IV: control paradigms and data structures},
  volume 1611, pages 586--606. International Society for Optics and Photonics,
  1992.

\bibitem[Bishop(1994)]{bishop1994mixture}
Christopher~M Bishop.
\newblock Mixture density networks.
\newblock 1994.

\bibitem[Chen et~al.(2019)Chen, Chen, Sui, Ye, Liu, Bahar, and
  Jenkins]{chen2019grip}
Xiaotong Chen, Rui Chen, Zhiqiang Sui, Zhefan Ye, Yanqi Liu, R~Iris Bahar, and
  Odest~Chadwicke Jenkins.
\newblock Grip: Generative robust inference and perception for semantic robot
  manipulation in adversarial environments.
\newblock In \emph{2019 IEEE/RSJ International Conference on Intelligent Robots
  and Systems (IROS)}, pages 3988--3995. IEEE, 2019.

\bibitem[Cohen and Welling(2016{\natexlab{a}})]{cohen2016group}
Taco Cohen and Max Welling.
\newblock Group equivariant convolutional networks.
\newblock In \emph{International conference on machine learning}, pages
  2990--2999. PMLR, 2016{\natexlab{a}}.

\bibitem[Cohen and Welling(2016{\natexlab{b}})]{cohen2016steerable}
Taco~S Cohen and Max Welling.
\newblock Steerable cnns.
\newblock \emph{arXiv preprint arXiv:1612.08498}, 2016{\natexlab{b}}.

\bibitem[Curtis et~al.(2021)Curtis, Fang, Kaelbling, Lozano-P{\'e}rez, and
  Garrett]{curtis2021long}
Aidan Curtis, Xiaolin Fang, Leslie~Pack Kaelbling, Tom{\'a}s Lozano-P{\'e}rez,
  and Caelan~Reed Garrett.
\newblock Long-horizon manipulation of unknown objects via task and motion
  planning with estimated affordances.
\newblock \emph{arXiv preprint arXiv:2108.04145}, 2021.

\bibitem[Deng et~al.(2020)Deng, Xiang, Mousavian, Eppner, Bretl, and
  Fox]{deng2020self}
Xinke Deng, Yu~Xiang, Arsalan Mousavian, Clemens Eppner, Timothy Bretl, and
  Dieter Fox.
\newblock Self-supervised 6d object pose estimation for robot manipulation.
\newblock In \emph{2020 IEEE International Conference on Robotics and
  Automation (ICRA)}, pages 3665--3671. IEEE, 2020.

\bibitem[Devin et~al.(2020)Devin, Rowghanian, Vigorito, Richards, and
  Rohanimanesh]{devin2020self}
Coline Devin, Payam Rowghanian, Chris Vigorito, Will Richards, and Khashayar
  Rohanimanesh.
\newblock Self-supervised goal-conditioned pick and place.
\newblock \emph{arXiv preprint arXiv:2008.11466}, 2020.

\bibitem[Gualtieri and Platt(2021)]{gualtieri2021robotic}
Marcus Gualtieri and Robert Platt.
\newblock Robotic pick-and-place with uncertain object instance segmentation
  and shape completion.
\newblock \emph{IEEE Robotics and Automation Letters}, 6\penalty0 (2):\penalty0
  1753--1760, 2021.

\bibitem[He et~al.(2016)He, Zhang, Ren, and Sun]{he2016deep}
Kaiming He, Xiangyu Zhang, Shaoqing Ren, and Jian Sun.
\newblock Deep residual learning for image recognition.
\newblock In \emph{Proceedings of the IEEE conference on computer vision and
  pattern recognition}, pages 770--778, 2016.

\bibitem[Hester et~al.(2018)Hester, Vecerik, Pietquin, Lanctot, Schaul, Piot,
  Horgan, Quan, Sendonaris, Osband, et~al.]{hester2018deep}
Todd Hester, Matej Vecerik, Olivier Pietquin, Marc Lanctot, Tom Schaul, Bilal
  Piot, Dan Horgan, John Quan, Andrew Sendonaris, Ian Osband, et~al.
\newblock Deep q-learning from demonstrations.
\newblock In \emph{Thirty-second AAAI conference on artificial intelligence},
  2018.

\bibitem[Huang et~al.(2021)Huang, Yang, and Platt]{huang2021gascn}
Haojie Huang, Ziyi Yang, and Robert Platt.
\newblock Gascn: Graph attention shape completion network.
\newblock In \emph{2021 International Conference on 3D Vision (3DV)}, pages
  1269--1278. IEEE, 2021.

\bibitem[Hussein et~al.(2017)Hussein, Gaber, Elyan, and
  Jayne]{hussein2017imitation}
Ahmed Hussein, Mohamed~Medhat Gaber, Eyad Elyan, and Chrisina Jayne.
\newblock Imitation learning: A survey of learning methods.
\newblock \emph{ACM Computing Surveys (CSUR)}, 50\penalty0 (2):\penalty0 1--35,
  2017.

\bibitem[Khansari et~al.(2020)Khansari, Kappler, Luo, Bingham, and
  Kalakrishnan]{khansari2020action}
Mohi Khansari, Daniel Kappler, Jianlan Luo, Jeff Bingham, and Mrinal
  Kalakrishnan.
\newblock Action image representation: Learning scalable deep grasping policies
  with zero real world data.
\newblock In \emph{2020 IEEE International Conference on Robotics and
  Automation (ICRA)}, pages 3597--3603. IEEE, 2020.

\bibitem[Kingma and Ba(2014)]{kingma2014adam}
Diederik~P Kingma and Jimmy Ba.
\newblock Adam: A method for stochastic optimization.
\newblock \emph{arXiv preprint arXiv:1412.6980}, 2014.

\bibitem[Levine et~al.(2016)Levine, Finn, Darrell, and Abbeel]{levine2016end}
Sergey Levine, Chelsea Finn, Trevor Darrell, and Pieter Abbeel.
\newblock End-to-end training of deep visuomotor policies.
\newblock \emph{The Journal of Machine Learning Research}, 17\penalty0
  (1):\penalty0 1334--1373, 2016.

\bibitem[Liu et~al.(2020)Liu, Jonschkowski, Angelova, and
  Konolige]{liu2020keypose}
Xingyu Liu, Rico Jonschkowski, Anelia Angelova, and Kurt Konolige.
\newblock Keypose: Multi-view 3d labeling and keypoint estimation for
  transparent objects.
\newblock In \emph{Proceedings of the IEEE/CVF conference on computer vision
  and pattern recognition}, pages 11602--11610, 2020.

\bibitem[Morrison et~al.(2018)Morrison, Corke, and
  Leitner]{morrison2018closing}
Douglas Morrison, Peter Corke, and J{\"u}rgen Leitner.
\newblock Closing the loop for robotic grasping: A real-time, generative grasp
  synthesis approach.
\newblock \emph{arXiv preprint arXiv:1804.05172}, 2018.

\bibitem[Nagabandi et~al.(2020)Nagabandi, Konolige, Levine, and
  Kumar]{nagabandi2020deep}
Anusha Nagabandi, Kurt Konolige, Sergey Levine, and Vikash Kumar.
\newblock Deep dynamics models for learning dexterous manipulation.
\newblock In \emph{Conference on Robot Learning}, pages 1101--1112. PMLR, 2020.

\bibitem[Nair and Hinton(2010)]{nair2010rectified}
Vinod Nair and Geoffrey~E Hinton.
\newblock Rectified linear units improve restricted boltzmann machines.
\newblock In \emph{Icml}, 2010.

\bibitem[Narayanan and Likhachev(2016)]{narayanan2016discriminatively}
Venkatraman Narayanan and Maxim Likhachev.
\newblock Discriminatively-guided deliberative perception for pose estimation
  of multiple 3d object instances.
\newblock In \emph{Robotics: Science and Systems}, 2016.

\bibitem[Qureshi et~al.(2021)Qureshi, Mousavian, Paxton, Yip, and
  Fox]{qureshi2021nerp}
Ahmed~H Qureshi, Arsalan Mousavian, Chris Paxton, Michael~C Yip, and Dieter
  Fox.
\newblock Nerp: Neural rearrangement planning for unknown objects.
\newblock \emph{arXiv preprint arXiv:2106.01352}, 2021.

\bibitem[Ronneberger et~al.(2015)Ronneberger, Fischer, and
  Brox]{ronneberger2015u}
Olaf Ronneberger, Philipp Fischer, and Thomas Brox.
\newblock U-net: Convolutional networks for biomedical image segmentation.
\newblock In \emph{International Conference on Medical image computing and
  computer-assisted intervention}, pages 234--241. Springer, 2015.

\bibitem[Serre(1977)]{serre1977linear}
Jean-Pierre Serre.
\newblock \emph{Linear representations of finite groups}, volume~42.
\newblock Springer, 1977.

\bibitem[Vecerik et~al.(2017)Vecerik, Hester, Scholz, Wang, Pietquin, Piot,
  Heess, Roth{\"o}rl, Lampe, and Riedmiller]{vecerik2017leveraging}
Mel Vecerik, Todd Hester, Jonathan Scholz, Fumin Wang, Olivier Pietquin, Bilal
  Piot, Nicolas Heess, Thomas Roth{\"o}rl, Thomas Lampe, and Martin Riedmiller.
\newblock Leveraging demonstrations for deep reinforcement learning on robotics
  problems with sparse rewards.
\newblock \emph{arXiv preprint arXiv:1707.08817}, 2017.

\bibitem[Wang et~al.(2022{\natexlab{a}})Wang, Walters, and
  Platt]{wang2022mathrm}
Dian Wang, Robin Walters, and Robert Platt.
\newblock {$\mathrm{SO}(2)$}-equivariant reinforcement learning.
\newblock In \emph{The Tenth International Conference on Learning
  Representations}, 2022{\natexlab{a}}.

\bibitem[Wang et~al.(2022{\natexlab{b}})Wang, Walters, Zhu, and
  Platt]{wang2022equivariant}
Dian Wang, Robin Walters, Xupeng Zhu, and Robert Platt.
\newblock Equivariant $ q $ learning in spatial action spaces.
\newblock In \emph{Conference on Robot Learning}, pages 1713--1723. PMLR,
  2022{\natexlab{b}}.

\bibitem[Weiler and Cesa(2019)]{weiler2019general}
Maurice Weiler and Gabriele Cesa.
\newblock General $ e (2) $-equivariant steerable cnns.
\newblock \emph{arXiv preprint arXiv:1911.08251}, 2019.

\bibitem[Yoon et~al.(2003)Yoon, DeSouza, and Kak]{yoon2003real}
Youngrock Yoon, Guilherme~N DeSouza, and Avinash~C Kak.
\newblock Real-time tracking and pose estimation for industrial objects using
  geometric features.
\newblock In \emph{2003 IEEE International Conference on Robotics and
  Automation (Cat. No. 03CH37422)}, volume~3, pages 3473--3478. IEEE, 2003.

\bibitem[Yuan et~al.(2018)Yuan, Khot, Held, Mertz, and Hebert]{yuan2018pcn}
Wentao Yuan, Tejas Khot, David Held, Christoph Mertz, and Martial Hebert.
\newblock Pcn: Point completion network.
\newblock In \emph{2018 International Conference on 3D Vision (3DV)}, pages
  728--737. IEEE, 2018.

\bibitem[Zakka et~al.(2020)Zakka, Zeng, Lee, and Song]{zakka2020form2fit}
Kevin Zakka, Andy Zeng, Johnny Lee, and Shuran Song.
\newblock Form2fit: Learning shape priors for generalizable assembly from
  disassembly.
\newblock In \emph{2020 IEEE International Conference on Robotics and
  Automation (ICRA)}, pages 9404--9410. IEEE, 2020.

\bibitem[Zeng et~al.(2018{\natexlab{a}})Zeng, Song, Welker, Lee, Rodriguez, and
  Funkhouser]{zeng2018learning}
Andy Zeng, Shuran Song, Stefan Welker, Johnny Lee, Alberto Rodriguez, and
  Thomas Funkhouser.
\newblock Learning synergies between pushing and grasping with self-supervised
  deep reinforcement learning.
\newblock In \emph{2018 IEEE/RSJ International Conference on Intelligent Robots
  and Systems (IROS)}, pages 4238--4245. IEEE, 2018{\natexlab{a}}.

\bibitem[Zeng et~al.(2018{\natexlab{b}})Zeng, Song, Yu, Donlon, Hogan, Bauza,
  Ma, Taylor, Liu, Romo, et~al.]{zeng2018robotic}
Andy Zeng, Shuran Song, Kuan-Ting Yu, Elliott Donlon, Francois~R Hogan, Maria
  Bauza, Daolin Ma, Orion Taylor, Melody Liu, Eudald Romo, et~al.
\newblock Robotic pick-and-place of novel objects in clutter with
  multi-affordance grasping and cross-domain image matching.
\newblock In \emph{2018 IEEE international conference on robotics and
  automation (ICRA)}, pages 3750--3757. IEEE, 2018{\natexlab{b}}.

\bibitem[Zeng et~al.(2020)Zeng, Florence, Tompson, Welker, Chien, Attarian,
  Armstrong, Krasin, Duong, Sindhwani, et~al.]{zeng2020transporter}
Andy Zeng, Pete Florence, Jonathan Tompson, Stefan Welker, Jonathan Chien,
  Maria Attarian, Travis Armstrong, Ivan Krasin, Dan Duong, Vikas Sindhwani,
  et~al.
\newblock Transporter networks: Rearranging the visual world for robotic
  manipulation.
\newblock \emph{arXiv preprint arXiv:2010.14406}, 2020.

\end{thebibliography}
\newpage \newpage
\clearpage
\section{Appendix}
\subsection{Equivariance under Proposition 2:}
\label{special_properties}
We summarize some important properties related to our place network which follow from Proposition 2 or its proof and provide a intuitive explanation for each one. Recall that Proposition 2 states:
\begin{equation*}
    \begin{aligned}
        %\psi(g \cdot Crop) \star \phi (-g \cdot O_t/Crop) =  \psi(-g \cdot Crop) \star \phi (g \cdot O_t/Crop)
        &\Psi(T^0_{g_1}(c)) \star \phi (T^0_{g_2} (o_t)) \\ &\qquad=\rho_{\reg}(g_2-g_1)(T^0_{g_2}[\Psi(c) \star \phi ( o_t)]).
        % f_\place(T^0_{g_1}(c), T^0_{g_2} (o_t\backslash c)) = \rho_{reg}(g_2-g_1)T^0_{g_2} f_\place( c, o_t\backslash c)
    \end{aligned}
    \label{generalized}
\end{equation*}
Then he have the following properties.

\paragraph{Equivariance property}
Setting $g_1=0$ or $g_2=0$ we get respectively
\begin{align}
    \Psi(T^0_g(c)) \star \phi (o_t) &= \rho_{\reg}(-g) (\Psi(c) \star \phi (o_t))
    \label{equi_original_our} \\
    \Psi(c) \star \phi (T^0_g(o_t)) &= T^{\reg}_g(\Psi(c) \star \phi (o_t))
    \label{equi_2_simple}
\end{align}
These show the equivariance of our network $f_\place$ under either a rotation $g\in C_n$ of the object or the placement.

\paragraph{Invariance property} Setting $g_1=g_2$, we get
\begin{equation*}
    \Psi(T^0_g(c)) \star \phi (T^0_g( o_t)) = T^0_g (\Psi(c) \star \phi (o_t)).
\end{equation*}
%clarify that the channel is invariant, the pixels are equivariant
This equation demonstrates that a rotation $g$ on the whole observation $o_t$ does not change the placing angle but rotates the placing location by $g$. Although data augmentation could help non-equivariant models learn this property, our networks observe it by construction.
% Motivate this from symmetry of problem (in introduction).

\paragraph{Relativity property}
Related to Equation \ref{equi_original_our}, we also have
\begin{equation*}
    \begin{aligned}
        %\psi(g \cdot Crop) \star \phi (-g \cdot O_t/Crop) =  \psi(-g \cdot Crop) \star \phi (g \cdot O_t/Crop)
        \Psi(T^0_g(c)) \star \phi (o_t) =  \rho_{reg}(-g)(T^0_g[ (\Psi(c) \star \phi ((T^0_{-g}( o_t))]).
    \end{aligned}
    \label{relativity}
\end{equation*}
This equation defines the relationship between a rotation on $c$ by $g$ and a 
inverse rotation $-g$ on $o_t$. %They are equivariant under some transformation.
Intuitively, $c$ could be considered as the L-shaped block and $o_t$ can be regarded as the L-shaped slot.

\subsection{Proofs of propositions}
\label{proof_proposition}
We now prove Proposition 1 and Proposition 2.  We start with some common lemmas. In order to understand continuous rotations of image data, it is helpful to consider a $k$-channel image as a mapping $f\colon \mathbb{R}^2 \mapsto \mathbb{R}^k$ where the input $\mathbb{R}^2$ defines the pixel space.  We consider images centered at $(0,0)$ and for non-integer values $(x,y)$ we consider $f(x,y)$ to be the interpolated pixel value.  Similarly, let $K: \mathbb{R}^2 \mapsto \mathbb{R}^{l\times k}$ be convolutional kernel where $k$ is the number of the input channels and $l$ is the number of the output channels.  Although the input space is $\mathbb{R}^2$, we assume the kernel is $r\!\times\!r$ pixels and $K(x,y)$ is zero outside this set.  The convolution can then be expressed by
  $(K\star f)(\vec{v})= \sum_{\vec{w}\in \mathbb{Z}^2}f(\vec{v}+\vec{w})K(\vec{w })$, where $\vec{v}=(i,j)\in\mathbb{R}^2$. 
  %%%%%%======================
  \\
  \begin{lemma}
  \begin{align}\label{eqn:lemma1}
    (T_g^0(K\star f))(\vec{v})= ((T_g^0 K) \star (T_g^0 f))(\vec{v})  
  \end{align}
  \end{lemma}
  \noindent
  \begin{proof}
   We evaluate the left hand side of Equation \ref{eqn:lemma1}.
  \begin{align*}
        T_g^0(K \star f)(\vec{v}) &=  \sum_{\vec{w}\in \mathbb{Z}^2}f(g^{-1}\vec{v}+\vec{w})K(\vec{w }).
  \end{align*}
  Re-indexing the sum with $\vec{y} = g \vec{w}$,
    \begin{align*}
        &= \sum_{\vec{y}\in \mathbb{Z}^2}f(g^{-1}\vec{v}+g^{-1}\vec{y})K(g^{-1}\vec{y })
    \end{align*}
    is by definition
    \begin{align*}
        &= \sum_{\vec{y}\in \mathbb{Z}^2}(T^0_gf)(\vec{v}+\vec{y})(T^0_gK)(\vec{y }) \\
        &= ((T_g^0K)\star (T_g^0f))(\vec{v})
    \end{align*}
    as desired.
  \end{proof}
  
 %Define $f\colon \mathbb{R}^2 \mapsto \mathbb{R}^k$ and
 Assume input $f \colon \mathbb{R}^2 \to \mathbb{R}$. Consider a diagonal kernel $\tilde{K} \colon  \mathbb{R}^2 \mapsto \mathbb{R}^{n\times n}$ where $\tilde{K}(\vec{v})$ is a diagonal $n \times n$ matrix $\mathrm{Diag}(K_1,\ldots,K_n)$.  Define $\tilde f \colon \mathbb{R}^2 \to \mathbb{R}^{n}$ to be the $n$-fold duplication of $f$ such that $\tilde{f}(\vec{v}) = (f(\vec{v}), \ldots, f(\vec{v}))$.   
% Define $\tilde{K}=\mathrm{Diag}(K_1,\ldots,K_n) \colon \mathbb{R}^2 \mapsto \mathbb{R}^{n\times n}$ and $\tilde{f} &= Dup(f) \colon \mathbb{R}^2 \mapsto \mathbb{R}^n$, where $\tilde{f}(\vec{x})= (f(\vec{x}),\ldots,f(\vec{x}))$. 
 For such inputs and kernels, we have the following permutation equivariance.
     
  \begin{lemma}\label{lem:convequ}   
\begin{align*}
    (\rho_{\mathrm{reg}}(g)\tilde{K}) \star \tilde{f} &= \rho_{\mathrm{reg}}(g)(\tilde{K}\star \tilde{f})
\end{align*}
   \end{lemma}
   
   \begin{proof}
   By definition $h_i = (\tilde{K}\star \tilde{f})_i = K_i \star f$. 
     Clearly permuting the 1x1 kernels $K_i$ also permutes $h_i$, so $\rho_{\mathrm{reg}}(g)h = (\rho_{\mathrm{reg}}(g)\tilde{K})\star \tilde{f}$ as desired.
   \end{proof}
   
   We require one more lemma on the equivariance of $\mathcal{R}_n$.
   
   \begin{lemma}\label{lem:rnequ}
   \begin{align*}
       \mathcal{R}_n(T_g^0 f) = \rho_{\mathrm{reg}}(-g)\mathcal{R}_n(f)
   \end{align*}
   \end{lemma}
   
   \begin{proof}
   First we compute 
     \begin{align*}
         \mathcal{R}_n(f)(\vec{x})&=( f(\vec{x}),  f(g^{-1}\vec{x}),\ldots ,f(g^{-(n-1)}\vec{x})).
    \end{align*}
    Then both $\mathcal{R}_n(T_g^0 f)$ and  $\rho_{\mathrm{reg}}(-g)\mathcal{R}_n(f)$ equal 
    \begin{align*}
        (f(g^{-1}\vec{x}),\ldots,f(g^{-(n-1)}\vec{x}), f(\vec{x})).
     \end{align*}
   \end{proof}
   
\subsubsection{Proof of Proposition 1}
We prove the equivariance of Transporter Net under rotations of the picked object,
  \begin{equation}
    \label{eqn:appprop1}
      \psi(\mathcal{R}_n(T^0_{g}c)) \star \phi(o_t) = \rho_{\reg}(-g)( \psi(\mathcal{R}_n(c)) \star \phi(o_t)
  \end{equation}
  \begin{proof}
   Since $\psi$ is applied independently to each of the rotated channels in $\mathcal{R}_n(c)$, we denote $\psi_n((f_1,\ldots,f_n))=(\psi(f_1),\ldots,(\psi(f_n))$.  By Lemma \ref{lem:rnequ}, the left-hand side of Equation \ref{eqn:appprop1} is  
  \begin{align*}
  \psi(\mathcal{R}_n(T^0_{g}c)) \star \phi(o_t) &= \psi_n(\rho_{\reg}(-g)\mathcal{R}_n(c)) \star \phi(o_t). 
  \end{align*}
  Since $\psi_n$ applies $\psi$ on each component, it is equivariant to permutation of components and thus the above becomes
  \begin{align*}    
       &= (\rho_{\reg}(-g) \psi_n(\mathcal{R}_n(c)) \star \phi(o_t).
  \end{align*}
  Finally applying Lemma \ref{lem:convequ} gives 
  \begin{align*}
      &= \rho_{\reg}(-g) (\psi_n(\mathcal{R}_n(c) \star \phi(o_t))
  \end{align*}
  as desired.
  \end{proof}

\subsubsection{Proof of Proposition 2}
Recall $\Psi(c) = \psi(\mathcal{R}_n(c))$. We now prove Proposition 2, 
\begin{equation}
      \Psi(T^0_{g_1}(c)) \star \phi (T^0_{g_2} (o_t\setminus c)) =\\ \rho_{reg}(g_2-g_1)(T^0_{g_2}[\Psi(c) \star \phi ( o_t\setminus c)])
      \label{equ:proof1}
\end{equation}

\begin{proof}
 We first first prove the equivariance under rotations of the placement $o_t$.  We claim
\begin{equation}\label{eqn:prop2subclaim}
    % \Psi(c) \star \phi (T^0_g(o_t)) = \rho_{reg}(g) (T^0_g( \Psi(c) \star \phi (o_t))),
    \Psi(c) \star \phi (T^0_g(o_t)) = T_g^{\reg}( \Psi(c) \star \phi (o_t)).
  \end{equation}
Evaluating the left hand side of Equation \ref{eqn:prop2subclaim},
 \begin{align*}
    &\Psi(c) \star \phi (T^0_g(o_t)) \\
     &\qquad= \Psi(c) \star T^0_g \phi (o_t) \:\:\:\: \text{(equivariance of $\phi$)}\\
     &\qquad= (T^0_g T^0_{g^{-1}}\Psi(c)) \star (T^0_g \phi (o_t))\\
     &\qquad= T^0_g (T^0_{g^{-1}}\Psi(c) \star \phi (o_t))\:\: \text{(Lemma 8.1)}\\
     &\qquad= T^0_g (T^0_{g^{-1}}\mathcal{R}_n(\psi (c)) \star \phi (o_t))\\
     &\qquad= T^0_g (\mathcal{R}_n(\psi (T^0_{g^{-1}} c)) \star \phi (o_t)) \:\: \text{(equiv. of $\psi$, $\mathcal{R}_n$)}\\
     &\qquad= T^0_g ((\rho_{\reg}(g)\Psi(c) \star \phi (o_t)) \:\:\text{(Lemma 8.3)}\\
     &\qquad= T^0_g \rho_{\reg}(g)(\Psi(c) \star \phi (o_t)) \:\: \text{(Lemma 8.2)}\\
     &\qquad=  T_g^{\reg}( \Psi(c) \star \phi (o_t)).
 \end{align*}
 In the last step $T_g^{\reg} = \rho_{\reg}(g) T^0_g  = T^0_g \rho_{\reg}(g)$ since  $T^0_g$ and $\rho_{\reg}(g)$ commute as $\rho_{\reg}(g)$ acts on channel space and $T_g^0$ acts on base space.  This proves the claim of Equation \ref{eqn:prop2subclaim}.
 
 Now, using the equivariance of $\psi$, Proposition 1 may be reformulated as
 \begin{equation}
      \Psi(T^0_{g}c) \star \phi(o_t\setminus c) = \rho_{\reg}(-g)( \Psi(c) \star) \phi(o_t \setminus c)
      \label{equ:proof2}
 \end{equation}
 Note we use $o_t\setminus c$ to emphasize the target placement since the object and the placement are non-overlapping. Combining \eref{equ:proof1} with \eref{equ:proof2} realizes the Proposition 2. 
\end{proof}

\subsection{Task descriptions of Ravens-10:}
\label{raven_10_details}
Here we provide a short description of Ravens-10 Environment, we refer readers to~\cite{zeng2020transporter} for details.
The poses of objects and placements in each task are randomly sampled in the workspace without collision. Performance on each task is evaluated in one of two ways: 1) pose: translation and rotation error relative to target pose is less than a threshold $ \tau=1\mathrm{cm}$ and $\omega=\frac{\pi}{12}$ respectively. Tasks: block-insertion, towers-of-hanoi, place-red-in-green, align-box-corner, stack-block-pyramid, assembling-kits. Partial scores are assigned to multiple-action tasks. 2) Zone:  Ravens-10 discretizes the 3D bounding box of each object into $2cm^3$ voxels. The Total reward is calculated by $\frac{\text{\# of voxels in target zone}}{\text{total \# of voxels}}$. Tasks: palletizing-boxes, packing-boxes, manipulating-cables, sweeping-piles.
Note that pushing objects could also be parameterized with $a_{\mathrm{pick}}$ and $a_\mathrm{{place}}$ that correspond to the starting pose and the ending pose of the end effector.
\begin{enumerate}
    \item \textbf{block-insertion:} pick up an L-shape block and place it into an L-shaped fixture. 
    \item \textbf{place-red-in-green:} picking up red cubes and place them into green bowls. There could be multiple bowls and cubes with different colors.
    \item \textbf{towers-of-hanoi:} sequentially picking up disks and placing them into pegs such that all 3 disks initialized on one peg are moved to another, and that only smaller disks can be on top of larger ones.
    \item \textbf{align-box-corner:} picking up a randomly sized box and place it to align one of its
    corners to a green L-shaped marker labeled on the tabletop.
    \item \textbf{stack-block-pyramid:} sequentially picking up 6 blocks and stacking them into a pyramid of 3-2-1.
    \item \textbf{palletizing-boxes:} picking up 18 boxes and stacking them on top of a pallet.
    \item \textbf{assembling-kits:} picking 5 shaped objects (randomly sampled with replacement from a set of 20) and fitting them to corresponding silhouettes of the objects on a board.
    \item \textbf{packing-boxes:} picking and placing randomly sized boxes tightly into a randomly sized container.
    \item \textbf{manipulating-rope:} manipulating a deformable rope such that it connects the two endpoints of an incomplete 3-sided square (colored in green).
    \item \textbf{sweeping-piles:} pushing piles of small objects (randomly initialized) into a desired target goal zone on the tabletop marked with green boundaries. The task is implemented with a pad-shaped end effector. 
\end{enumerate}

\end{document}

\maketitle
  %starter
  \noindent
  \section*{Proof}
  Define $f\colon \mathbb{R}^2 \mapsto \mathbb{R}^k$ and $K: \mathbb{R}^2 \mapsto \mathbb{R}^{k\times l}$, where $k$ is the number of the input channels and $l$ is the number of the output channel. The convolution can be expressed by
  $(K\ast f)(\vec{v})= \sum_{\vec{w}}f(\vec{v}+\vec{w})K(\vec{w})$, where $\vec{v}=(i,j)\in\mathbb{R}^2$.
  
%%%%%%================================================================================  
  \begin{Claim}
   $(T_g^0(K\ast f))(\vec{v})= (T_g^0 K) \ast (T_g^0 f)(\vec{v})$
  \end{Claim}
  \begin{proof}\\
  $ RHS&=\sum_{\vec{w}} f(g^{-1}(\vec{v}+\vec{w}))K(g^{-1}\vec{w})$; \\ \\
  Let $\vec{y}=g^{-1}\vec{w}$, $LHS&=  (K \ast f)(g^{-1}\vec{v})= \sum_{\vec{y}} f(g^{-1}\vec{v} +\vec{y})K(\vec{y}) = \sum_{\vec{w}}f((g^{-1}\vec{v} +g^{-1}\vec{w})K(g^{-1}\vec{w})$\\ \\
  $RHS \Leftrightarrow LHS$
  \end{proof}
%%%%%%================================================================================  
  \begin{Claim}
     $f\colon \mathbb{R}^2 \mapsto \mathbb{R}^k$ and $K_{i=1\colon n} \colon  \mathbb{R}^2 \mapsto \mathbb{R}^{1\times1}$.
     $\tilde{K}=\mathrm{Diag}(K_1,\ldots,K_n) \colon \mathbb{R}^2 \mapsto \mathbb{R}^{n\times n}$. $\tilde{f} &= Dup(f) \colon \mathbb{R}^2 \mapsto \mathbb{R}^n$, where $\tilde{f}(\vec{x})= (f(\vec{x}),\ldots,f(\vec{x}))$. So $(\tilde{K}\ast \tilde{f})_i &= K_i \ast f$. For $g\in C_n$, $(\rho_{\mathrm{reg}}(g)\tilde{K}) \ast \tilde{f} &= \rho_{\mathrm{reg}}(g)(\tilde{K}\ast \tilde{f})$.
   \end{Claim}
   
  \begin{proof}
   Assume $n$ kernels $K_{i&=1\colon n}$, then we convolve each kernel with the input image $h_i=(K_i\ast f)$. Clearly permuting the kernels $K_i$ also permutes $h_i$, so $\rho_{\mathrm{reg}}(g)h = (\rho_{\mathrm{reg}}(g)K)\ast f$
  \end{proof}
  
 %%%%%%================================================================================  
  \begin{Claim}
     $\mathcal{R}_n(T_g^0 f) = \rho_{\mathrm{reg}}(-g)\mathcal{R}_n(f) $
  \end{Claim}
  \begin{proof}
   $\mathcal{R}_n(f)(\vec{x})=( f(\vec{x}),  f(g^{-1}\vec{x}), f(g^{-2}\vec{x}),\ldots ,f(g^{-(n-1)}\vec{x}))$;\\ \\
   $RHS &= LHS = (f(g^{-1}\vec{x}), f(g^{-2}\vec{x}),\ldots,f(g^{-(n-1)}\vec{x}), f(\vec{x}))$
  \end{proof}  
%%%%%%================================================================================  
\subsection{Proof of proposition 1}
The equivariance of Transporter Net under rotations of the picked object:
%   \begin{equation*}
%       f_{\place}(o_t,c) = \psi(\mathcal{R}_n(c)) \star \phi(o_t).
%   \end{equation*}
%   \begin{equation*}
%         \label{eqn:transporter}
%         f_{\place}(o_t,T^0_g(c))  = \rho_{reg}(-g) f_{\place}(o_t,c).
%   \end{equation*}
  \begin{equation}
      \psi(\mathcal{R}_n(T^0_{g}c)) \star \phi(o_t) = \rho_{\reg}(-g)( \psi(\mathcal{R}_n(c)) \star) \phi(o_t)
  \end{equation}
  
  \begin{proof}
   Since $\psi$ is applied independently to each of the rotated channels in $\mathcal{R}_n(c)$, we can define $\psi_n((f_1,\ldots,f_n))=(\psi(f_1),\ldots,(\psi(f_n))$. 
   \begin{align*}
   LHS &= \psi_n(\rho_{\reg}(g)\mathcal{R}_n(c)) \star \phi(o_t)\\
       &= (\rho_{\reg}(-g) \psi_n(\mathcal{R}_n(c)) \star \phi(o_t) \:\: {\text{claim 0.3}}\\
       &= \rho_{\reg}(-g) (\psi_n(\mathcal{R}_n(c) \star \phi(o_t)) \:\: {\text{claim 0.2}}\\
       & = RHS
   \end{align*}
  \end{proof}

\subsection{Proof of proposition 2}
To prove the equivariance of Equivariant Transporter Net under rotations of the picked object and the placement:

\begin{equation}
      \Psi(T^0_{g_1}(c)) \star \phi (T^0_{g_2} (o_t\setminus c)) =\\ \rho_{reg}(g_2-g_1)(T^0_{g_2}[\Psi(c) \star \phi ( o_t\setminus c)])
      \label{equ:proof1}
  \end{equation}
%%%%%%================================================================================    
 We need first prove the equivariance under rotations of the placement:
  \begin{equation}
    % \Psi(c) \star \phi (T^0_g(o_t)) = \rho_{reg}(g) (T^0_g( \Psi(c) \star \phi (o_t))),
    \Psi(c) \star \phi (T^0_g(o_t\setminus c)) = T_g^{\reg}( \Psi(c) \star \phi (o_t\setminus c)).\\
  \end{equation}
 
\begin{proof}
 \begin{align*}
     LHS &= \Psi(c) \star T^0_g \phi (o_t) \:\:\text{the equivariance of $\phi$}\\
     &= T^0_g T^0_{g^-1}\Psi(c) \star T^0_g \phi (o_t)\\
     &= T^0_g (T^0_{g^-1}\Psi(c) \star \phi (o_t))\:\: \text{claim 0.1}\\
     &= T^0_g (T^0_{g^-1}\mathcal{R}_n(\psi (c)) \star \phi (o_t))\\
     &= T^0_g ((\rho_{\reg}(g)\Psi(c) \star \phi (o_t)) \:\:\text{claim 0.3}\\
     &= T^0_g \rho_{\reg}(g)(\Psi(c) \star \phi (o_t)) \:\:\text{commute since one acts on channel space and one acts on base space}\\
     &=  T_g^{\reg}( \Psi(c) \star \phi (o_t))\\
 \end{align*}
 With the equivariance of $\psi$, proposition 1 could be reformulated as
 \begin{equation}
      \Psi(T^0_{g}c) \star \phi(o_t\setminus c) = \rho_{\reg}(-g)( \Psi(c) \star) \phi(o_t \setminus c)
      \label{equ:proof2}
 \end{equation}
 Note we use $o_t\setminus c$ to emphasize the target placement of $o_t$ since the object and the placement are nonoverlapping. Combining euq1 and equ2 realizes the proposition 2. 
\end{proof}

\section*{Special Equivariance of Proposition 2:}

In this subsection, we make a summary of some important properties inside our placing networks and also provide a intuitive explanation for each one. \\\\
\textbf{Equivariance property:}
When $g_1=0$ or $g_2=0$, we can get
\begin{equation*}
    \Psi(T^0_g(c)) \star \phi (o_t \backslash c) = \rho_{reg}(-g) (\Psi(c) \star \phi (o_t\backslash c))
    \label{equi_original_our}
\end{equation*}

\begin{equation*}
    \Psi(c) \star \phi (T^0_g(o_t \backslash c)) = T^{reg}_g(\Psi(c) \star \phi (o_t\backslash c))
    \label{equi_2_simple}
\end{equation*}
Both of and show the equivariance of our networks under the rotation $g\in C_n$ of either the object or the placement.
\\\\
\textbf{Invariance property:} when $g_1=g_2$,
\begin{equation}
    \Psi(T^0_g(c)) \star \phi (T^0_g( o_t\backslash c)) = T^0_g (\Psi(c) \star \phi (o_t\backslash c))
\end{equation}
%clarify that the channel is invariant, the pixels are equivariant
The equation above demonstrates that a rotation $g$ on the whole observation $o_t$ doesn't change the placing angle but rotates the placing location by $g$. Although data augmentation could help non-equivairant models learn this property, our networks obtain it by nature.
% Motivate this from symmetry of problem (in introduction).
\\\\
\textbf{Relativity property:}\\
\begin{equation}
    \begin{aligned}
        %\psi(g \cdot Crop) \star \phi (-g \cdot O_t/Crop) =  \psi(-g \cdot Crop) \star \phi (g \cdot O_t/Crop)
        \Psi(T^0_g(c)) \star \phi (o_t\backslash c) =  \rho_{reg}(-g)(T^0_g[ (\Psi(c) \star \phi ((T^0_{-g}( o_t\backslash c))])
    \end{aligned}
    \label{relativity}
\end{equation}
% \begin{equation}
%     \begin{aligned}
%         %\psi(g \cdot Crop) \star \phi (-g \cdot O_t/Crop) =  \psi(-g \cdot Crop) \star \phi (g \cdot O_t/Crop)
%         \Psi(T^0_g(c_{im})) \star \phi (o_t\backslash c_{im}) =  \rho_{reg}(-g)(\\\rho_0(g)\cdot(\Psi(c_{im}) \star \phi (\rho_0(-g) \cdot o_t\backslash c_{im})))
%     \end{aligned}
%     \label{relativity}
% \end{equation}
\eref{relativity} explores the relationship between a rotation on $c$ by $g$ and a inverse rotation $-g$ on $o_t\backslash c$. They are equivariant under some transformation. Intuitively, $c$ could be considered as the block and $o_t\backslash c$ can be regarded as the slot.

\newpage
\section*{Task descriptions of Ravens-10:}
Here we provide a short description of Ravens-10 Environment, we refer readers to~\cite{zeng2020transporter} for details.
The poses of objects and placements in each task are randomly sampled in the workspace without collision. Performance on each task is evaluated in one of two ways: 1) pose: translation and rotation error relative to target pose is less than a threshold $ \tau=1\mathrm{cm}$ and $\omega=\frac{\pi}{12}$ respectively. Tasks: block-insertion, towers-of-hanoi, place-red-in-green, align-box-corner, stack-block-pyramid, assembling-kits. Partial scores are assigned to multiple-action tasks. 2) Zone:  Ravens-10 discretizes the 3D bounding box of each object into $2cm^3$ voxels. The Total reward is calculated by $\frac{\text{\# of voxels in target zone}}{\text{total \# of voxels}}$. Tasks: palletizing-boxes, packing-boxes, manipulating-cables, sweeping-piles.
Note that pushing objects could also be parameterized with $a_{\mathrm{pick}}$ and $a_\mathrm{{place}}$ that correspond to the starting pose and the ending pose of the end effector.
\begin{enumerate}
    \item \textbf{block-insertion:} pick up an L-shape block and place it into an L-shaped fixture. 
    \item \textbf{place-red-in-green:} picking up red cubes and place them into green bowls. There could be multiple bowls and cubes with different colors.
    \item \textbf{towers-of-hanoi:} sequentially picking up disks and placing them into pegs such that all 3 disks initialized on one peg are moved to another, and that only smaller disks can be on top of larger ones.
    \item \textbf{align-box-corner:} picking up a randomly sized box and place it to align one of its
    corners to a green L-shaped marker labeled on the tabletop.
    \item \textbf{stack-block-pyramid:} sequentially picking up 6 blocks and stacking them into a pyramid of 3-2-1.
    \item \textbf{palletizing-boxes:} picking up 18 boxes and stacking them on top of a pallet.
    \item \textbf{assembling-kits:} picking 5 shaped objects (randomly sampled with replacement from a set of 20) and fitting them to corresponding silhouettes of the objects on a board.
    \item \textbf{packing-boxes:} picking and placing randomly sized boxes tightly into a randomly sized container.
    \item \textbf{manipulating-rope:} manipulating a deformable rope such that it connects the two endpoints of an incomplete 3-sided square (colored in green).
    \item \textbf{sweeping-piles:} pushing piles of small objects (randomly initialized) into a desired target goal zone on the tabletop marked with green boundaries. The task is implemented with a pad-shaped end effector. 
\end{enumerate}

\section*{Training settings:}
We train Equivariant Transporter Network with Adam~\cite{} optimizer with a fixed learning rate of $10^{-4}$. It takes about 0.8 seconds to complete one SGD~\cite{} step with a batch size of 1 on a NVIDIA Tesla V100 SXM2 GPU. We evaluate the performance every 10k steps on 100 unseen tests for each task and best performances on most tasks can be achieved before 10K steps. 

\section*{Ablation study:}